\documentclass{IEEEtran}

\usepackage{cite}
\usepackage[pdftex]{graphicx}

\usepackage{subcaption}
\usepackage{float}
\usepackage{array}
\newcolumntype{C}[1]{>{\centering\let\newline\\\arraybackslash\hspace{0pt}}m{#1}}

\usepackage{amsmath}
\usepackage{adjustbox}
\usepackage{algorithm}
\usepackage{algpseudocode}

\begin{document}

\title{Artificial Neural Microcircuits as Building Blocks: Concept and Challenges}

\author{
    Andrew Walter$^{1}$, 
    Shimeng Wu$^{1}$,
    Andy M. Tyrrell$^{1}$,
    Liam McDaid$^{2}$,
    Malachy McElholm$^{2}$,
    Nidhin Thandassery Sumithran$^{2}$,
    Jim Harkin$^{2}$,
    Martin A. Trefzer$^{1}$
\thanks{$^{1}$ School of Physics, Engineering \& Technology; University of York
	
$^{2}$ School of Computing, Engineering and Intelligent Systems, Ulster University}
}

\maketitle

\begin{abstract}
Artificial Neural Networks (ANNs) are one of the most widely employed forms of bio-inspired computation. However the current trend is for ANNs to be structurally homogeneous. Furthermore, this structural homogeneity requires the application of complex training \& learning tools that produce application specific ANNs, susceptible to pitfalls such as overfitting. In this paper, an new approach is explored, inspired by the role played in biology by Neural Microcircuits, the so called ``fundamental processing elements'' of organic nervous systems. How large neural networks, particularly Spiking Neural Networks (SNNs) can be assembled using Artificial Neural Microcircuits (ANMs), intended as off-the-shelf components, is articulated; the results of initial work to produce a catalogue of such Microcircuits though the use of Novelty Search is shown; followed by efforts to expand upon this initial work, including a discussion of challenges uncovered during these efforts \& explorations of methods by which they might be overcome.
\end{abstract}

\IEEEkeywords{Artificial Neural Networks, Spiking Neural Networks, Neural Microcircuits, Neuromorphic Systems, Novelty Search, Evolutionary Algorithms, Spike Train Distance Measurement}


\section{Introduction}

Artificial Neural Networks (ANNs), in their various forms, have come to be the backbone of many non-standard computational methods \cite{Prieto-2016}; with the architectures ranging from the heavily bio-mimetic Spiking Neural Networks (SNNs) \cite{Yamazaki-2022}, to the various technologies encompased by the umbreller of Deep Learning \cite{Alzubaidi-2021}. Generally speaking however, the methodologies employed to design all ANN variations can be divided into two broad categories: large, topologically homogeneous feed forward (figure 1a) \& recurrent networks (figure 1b) whose connection weights are adjusted by either increasingly complex machine learning \cite{Schmidhuber-2015} \cite{wang-2020} or genetic algorithms \cite{Shifei-2013}; or iterative evolutionary methodologies designed to produce complex topologies from scratch \cite{Stanley-2002} (figure 1c). With these approaches, the resultant networks are often application specific, which both limits flexibility and raises the specter of issues such as overfitting as a consequence of the limited breadth of training data available.

\begin{figure}[h]
    \centering
    \includegraphics[width=0.9\columnwidth]{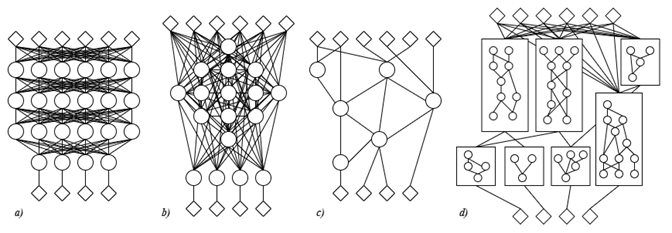}
    \caption{An illustration of the various approaches to neural network topologies, including the approach proposed in this paper: a) Feed Forward; b) Recurrent; c) Evolved from Scratch; d) Microcircuit based.}
    \label{fig:ANN_Topologies}
\end{figure}

However, advances in neuroscience are painting a picture of biological nervous systems that possess much more nuanced architectures; ones built up of various computational subunits, or Neural Microcircuits, which work together both hierarchically and in parallel \cite{Luo-2016}. Furthermore, these biological systems are more flexible, being able to carry out multiple different tasks, adapt to new ones, or apply approximate information to new situations.

In this paper, this more biological partitioned architecture is used as inspiration, employing novelty search to create a "component catalogue" of Artificial Neural Microcircuits, which can then be used as off-the-shelf components to fashion larger application specific networks, an idea illustrated in figure~\ref{fig:ANN_Topologies}d. It is envisaged that this approach will improve the robustness of network behaviours by breaking that behaviour into sub-behaviours that can be the focus of individual specialised Microcircuits or groups of Microcircuits; as well as bringing other advantages such as reducing network development and training overheads by allowing them to be built out of off-the-shelf components and allowing for the updating or alteration of a network’s overall behaviour through swapping of Microcircuits.

Section two will begin by detailing the concept of biological neural microcircuits, with particular reference to the neurology of the Signal Crayfish as an example organism. Section three will outline the methodology for producing a catalogue of Artificial Neural Microcircuits through the utilisation of Novelty Search, as well as articulating the proposal for employing so call neural Motifs as the fundamental building blocks. Section four outlines the methodology and results of an initial proof-of-concept experiment, where a rudimentary catalogue of ANMs was produced; before section five details the work done to refine the methodology of section three and produce a more comprehensive microcircuit catalogue. Section six explores the shortcomings exposed by the experiments of section five and the work done to investigate and ameliorate the causes of those shortcomings. Section seven then lays out some proposed paths forward for this work, articulating a few methods by which the ANM concept might be progressed further, before section eight concludes this work.    

\section{Neural Microcircuits}
Neuroscientists refer to the multitude of specialised subunits within biological nervous systems as Neural Microcircuits. Commonly referenced examples of such include motor neuron sequencing circuits; the hazard amelioration reflexes of some invertebrates; sensory processing circuits; and the columns of the neocortex. Though varied in form and function, Neural Microcircuits share a common role as the “elementary processing units” of the nervous system	\cite{Grillner-2006}.

\begin{figure}[h]
    \centering
    \includegraphics[width=0.9\columnwidth]{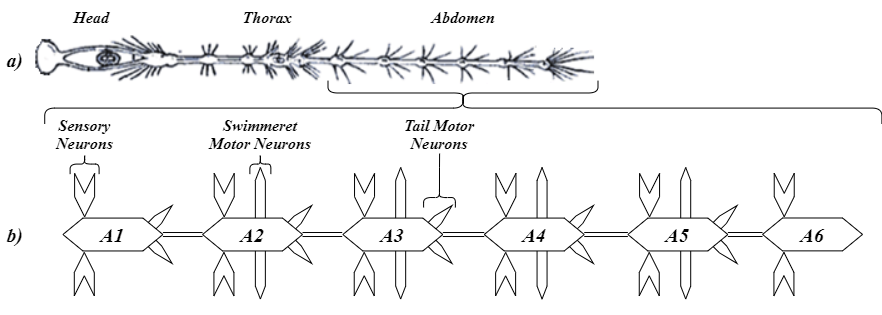}
    \caption{The crayfish ventral nerve cord (a), taken from \cite{Demyanenko-2019}; with a block illustration of the ganglia of the abdominal section (b), adapted from \cite{Smarandache-Wellmann-2014}}
    \label{fig:Crayfish_Nerve_Cord}
\end{figure}

By way of an example, consider two elements of the locomotion system of P. leniusculus, the Signal crayfish. This organism’s primary form of propulsion through the water is via a set of appendages called Swimmerets, which beat back and forth in coordination with one another to produce motion. Control of these appendages is performed by a portion of the crayfish’s nervous system called the Swimmer System, found within four of the six ganglia of the abdominal nerve cord. Each of these ganglia, labeled as A2 through A5 in figure~\ref{fig:Crayfish_Nerve_Cord}, controls the motion of one pair of Swimmerets via motor neurons RS \& PS.

\begin{figure}[h]
    \centering
    \includegraphics[width=0.7\columnwidth]{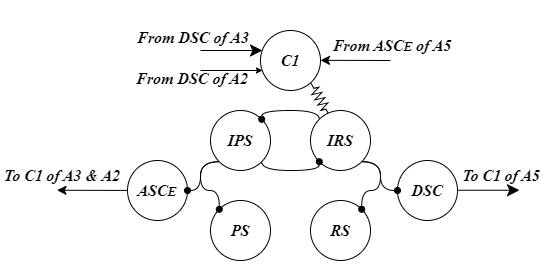}
    \caption{Illustration of the Microcircuit within the left hemiganglion of segment A4, counterparts of which are repeated in the segments A2 through A5. Arrows are connections to other microcircuits, dot lines are biochemical synapsis, \& the resistor represents an electrical synapsis. Adapted from \cite{Schneider-2018}}
    \label{fig:Crayfish_Hemiganglion_A4}
\end{figure}

Within each of the relevant hemiganglion is an identical neural Microcircuit, illustrated in figure~\ref{fig:Crayfish_Hemiganglion_A4}, consisting of a pair of motor neurons and associated inhibitory neurons: Power Stroke (PS), Inhibits Power Stroke (IPS), Return Stroke (RS) \& Inhibits Return Stroke (IRS); and three neurons that facilitate the interconnection of the Microcircuits in the different ganglia. In isolation, the pattern generating and motor neurons produce the rhythmic beating motion of the Swimmerets; but via the three interconnection neurons that motion is modulated, such that all the Swimmerets move in concert and thus efficient movement results \cite{Smarandache-Wellmann-2014}\cite{Schneider-2018}.

However, in situations where the crayfish is threatened, other locomotive behaviours are employed to facilitate fast escape. Collectively referred to Escape Reflexes, they activate the musculature that flexes the whole of the crayfish’s tail to rapidly propel the creature away from danger. An illustrative example of such an element is the Lateral Giant Escape reflex, named because of the role played by the Lateral Giant neurons within the nerve cord.

\begin{figure}[h]
    \centering
    \includegraphics[width=0.65\columnwidth]{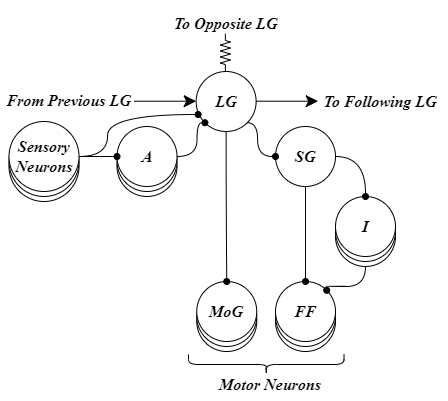}
    \caption{Illustration of the Lateral Giant Escape reflex Microcircuits within the left hemiganglion of segment A2, counterparts of which are repeated in segments A1 through A5. Adapted from \cite{Vu-1997}}
    \label{fig:Crayfish_Lateral_Giant}
\end{figure}

When stimulus is applied to the sensory neurons on the crayfish’s abdomen, signals are sent to the Lateral Giant (LG) neuron of that segment, either directly or through Interneuron A. As the LG neurons are connected in series down the abdomen and also linked to their counterparts on the opposite side of the animal, sufficient sensory input passes the threshold which is deemed to mean hostile action. The LG neuron then in turn activates the muscles of the tail, causing them to rapidly contract and push the crayfish out of danger. This motor action is caused by the LG directly signaling the Motor Giant neuron (MoG) in each segment, for the immediate motion, and the segments Segmental Giant neuron (SG), with in turn triggers the Fast Flexor neurons (FF) and their interneurons I, which extend the motion beyond the initial impulse \cite{Vu-1997}\cite{Edwards-2017}.

\subsection{Artificial Neural Microcircuits}
From the example detailed above, it is easy to see an argument for adopting a similar approach in neuromorphic systems; such compartmentalization of different functionality, but with interconnection to allow for coordination of more macro behaviours is seen as a positive side effect if it emerges within neural networks during training or evolution. Furthermore, as individual neural Microcircuits’ can be repeated throughout a nervous system wherever similar behaviours are needed, this paper proposes that it should be possible to create a general-purpose "component catalogue" of Artificial Neural Microcircuits, which can be used as off-the-shelf building blocks of a wide variety of different more complex neuromorphic systems, while assisting in scalable hardware implementations.

\section{Methodology}
In order to assemble a catalogue of Microcircuits that can be used, the following steps are required: (i) generation of candidate Microcircuits; (ii) assessment of them based on some criteria; and (iii) the use of the assessment to select which Microcircuits would be useful components.

\subsection{Generating Prospective Microcircuits}
A Microcircuit is represented using a set of connection matrices: (i) one for external inputs; (ii) one for internal connections; and (iii) a final array for outputs. This approach would allow for the production of arbitrary topologies simply by randomising the values of these matrices, however in this paper a somewhat different direction has been taken.

One of the goals of the proposed Microcircuits-as-components method is to provide a degree of substrate agnosticism. That is to say, the exact spiking neuron model that is used to implement the Microcircuits should be divorced from their functionality. Hence, it is desirable to avoid using individual neurons as the building blocks, and instead employ something “one step up” from this.

This can be achieved while also injecting some biological domain knowledge into the methodology through the use of neural Circuit Motifs. Neurobiologist Liqun Luo identifies ten recurring patterns of connectivity within biological neurocircuits, termed Motifs \cite{Luo-2016}. Each of these motifs consists of two to seven neurons and they exhibit distinct patterns of behaviour.

Initial investigations of the Microcircuit approach employ simplified versions of six of the ten motifs \cite{Byrn-1997}, selected because they represent a breadth of functionalities without using a large number of neurons and connections: FeedForward Excitation (FFE); FeedBack Excitation (FBE); FeedBack Inhibition (FBI); ReCurrent Excitation (RCE); ReCurrent Inhibition (RCI); and LaTeral Inhibition (LTI).

\begin{figure}[H]
    \centering
    \includegraphics[width=0.5\columnwidth]{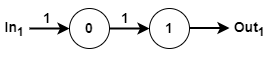}
    \caption{The FeedForward Excitation (FFE) Motif.}
    \label{fig:FFE}
\end{figure}

\begin{figure}[H]
    \centering
    \begin{subfigure}[t]{0.45\columnwidth}
        \includegraphics[width=1\linewidth]{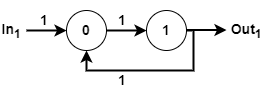} 
        \caption{}
        \label{fig:FBE}
    \end{subfigure}
    \begin{subfigure}[t]{0.45\columnwidth}
        \includegraphics[width=1\linewidth]{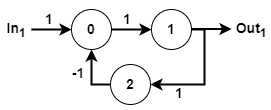} 
        \caption{}
        \label{fig:FBI}
    \end{subfigure}
    \caption{The (a) FeedBack Excitation (FBE) \& (b) FeedBack Inhibition (FBI) Motifs}
    \label{fig:FBE_FBI}
\end{figure}

\begin{figure}[H]
    \centering
    \begin{subfigure}[t]{0.45\columnwidth}
        \includegraphics[width=1\linewidth]{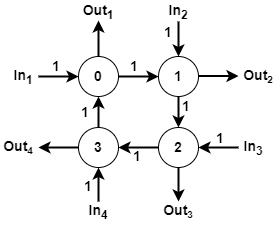} 
        \caption{}
        \label{fig:RCE}
    \end{subfigure}
    \begin{subfigure}[t]{0.45\columnwidth}
        \includegraphics[width=1\linewidth]{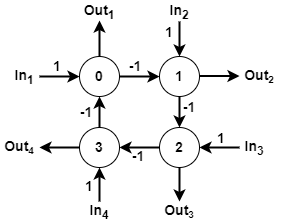} 
        \caption{}
        \label{fig:RCI}
    \end{subfigure}
    \caption{The (a) ReCurrent Excitation (RCE) \& (b) ReCurrent Inhibition (RCI) Motifs}
    \label{fig:RCE_RCI}
\end{figure}

\begin{figure}[H]
    \centering
    \includegraphics[width=0.4\linewidth]{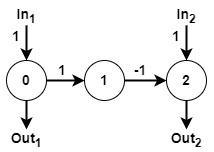}
    \caption{The LaTeral Inhibition (LTI) Motif}
    \label{fig:LTI}
\end{figure}

In addition to these, an additional “motif” is suggested, in the form of a Central Pattern Generator (CPG). This could be argued to be a Microcircuit in its own right, but here it is included as a “motif” due to its small size (three neurons), and the useful behaviour it adds (the ability to generate a consistently spiking output of a fixed frequency).

\begin{figure}[H]
    \centering
    \includegraphics[width=0.5\linewidth]{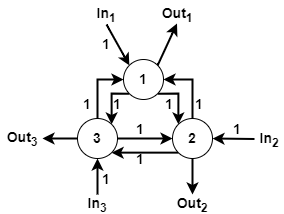}
    \caption{The Central Pattern Generator (CPG) Motif}
    \label{fig:CPG}
\end{figure}

\subsection{Microcircuit Connection Matrices}

To allow for the Motifs to be readily assembled into Microcircuits, a matrix-based approach has been adopted. Each Motif is described by a n-by-n matrix, where n is number of neurons in the Motif. These Connection Matrices can then be arbitrarily tiled together to layout the broad structure of a Microcircuit, onto which connections between Motifs can be added.

\begin{figure}[H]
    \centering
    \includegraphics[width=0.9\linewidth]{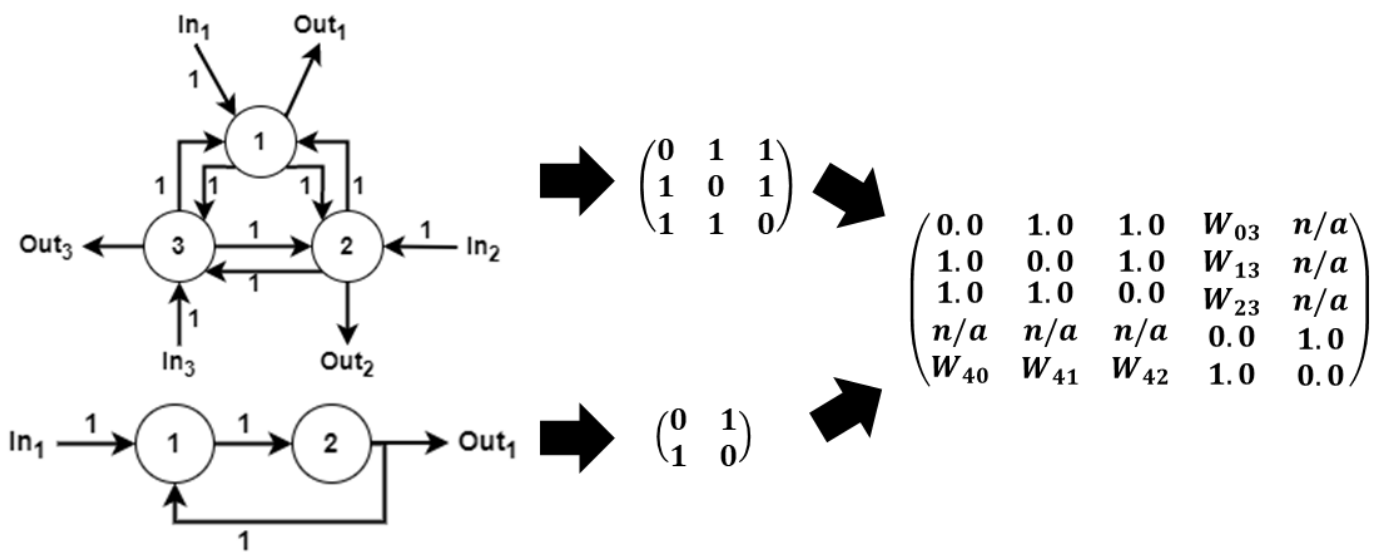}
    \caption{example of two Motifs (a CPG and FBE), their respective Connection Matrices, and a Microcircuit Connection Matrix formed by joining them together.}
    \label{fig:Matrix_Tiling_Example}
\end{figure}

Note that in the example in figure~\ref{fig:Matrix_Tiling_Example} a number of the possible connections between the two motifs are marked as n/a. This is because it is important to maintain the specific designation of neurons within the Motifs as inputs/outputs. In figure~\ref{fig:Matrix_Tiling_Example}, it can be seen that neuron 1 of the FBE can be the start of a connection as it’s an input; while neuron 2 of the FBE cannot be the end.

\begin{table*}[bth]
    \centering
    \begin{tabular}{|c|c|c|c|}
         \noalign{\hrule height 1pt}
        Motif & Connection Matrix & Input Neurons & Output Neurons \\
         \noalign{\hrule height 1.5pt}
        \shortstack{FeedForward Excitation\\ (FFE)} & \centering \begin{adjustbox}{valign=m} $ \begin{bmatrix}
                            0 & 1 \\
                            0 & 0
                        \end{bmatrix} $ \end{adjustbox} & 0 & 1 \\
         \noalign{\hrule height 1pt}
       \shortstack{FeedBack Excitation\\ (FBE)} &  \centering \begin{adjustbox}{valign=m}  $ \begin{bmatrix}
                            0 & 1 \\
                            1 & 0
                        \end{bmatrix} $ \end{adjustbox} & 0 & 1 \\
         \noalign{\hrule height 1pt}
        \shortstack{FeedBack Inhibition\\ (FBI)} &  \centering \begin{adjustbox}{valign=m}  $ \begin{bmatrix}
                            0 & 1 & 0 \\
                            0 & 0 & 1 \\
                            -1 & 0 & 0
                         \end{bmatrix} $ \end{adjustbox} & 0 & 2 \\
         \noalign{\hrule height 1pt}
        \shortstack{ReCurrent Excitation\\ (RCE)} & \centering \begin{adjustbox}{valign=m}  $ \begin{bmatrix}
                            0 & 1 & 0 & 0 \\
                            0 & 0 & 1 & 0 \\
                            0 & 0 & 0 & 1 \\
                            1 & 0 & 0 & 0
                        \end{bmatrix} $ \end{adjustbox} & 0,1,2,3 & 0,1,2,3 \\
        \noalign{\hrule height 1pt}
        \shortstack{ReCurrent Inhibition\\ (RCI)} & \centering \begin{adjustbox}{valign=m}  $ \begin{bmatrix}
                            0 & -1 & 0 & 0 \\
                            0 & 0 & -1 & 0 \\
                            0 & 0 & 0 & -1 \\
                            -1 & 0 & 0 & 0
                        \end{bmatrix} $ \end{adjustbox} & 0,1,2,3 & 0,1,2,3 \\
        \noalign{\hrule height 1pt}
        \shortstack{LaTeral Inhibition\\ (LTI)} &  \centering \begin{adjustbox}{valign=m}  $ \begin{bmatrix}
                            0 & 1 & 0 \\
                            0 & 0 & -1 \\
                            0 & 0 & 0
                         \end{bmatrix} $ \end{adjustbox} & 0,2 & 0,2 \\ 
        \noalign{\hrule height 1pt}
       \shortstack{Centeral Pattern\\ Generator (CPG)} &  \centering \begin{adjustbox}{valign=m}  $ \begin{bmatrix}
                            0 & 1 & 1 \\
                            1 & 0 & 1 \\
                            1 & 1 & 0
                         \end{bmatrix} $ \end{adjustbox} & 0,1,2 & 0,1,2 \\
        \noalign{\hrule height 1pt}
    \end{tabular}
    \caption{\label{tab:table-name} Connection Matrices and valid Input/Output neurons for all Motifs}
    \label{table:Motif_Matrices}
\end{table*}

\begin{figure}[H]
    \centering
    \includegraphics[width=0.9\linewidth]{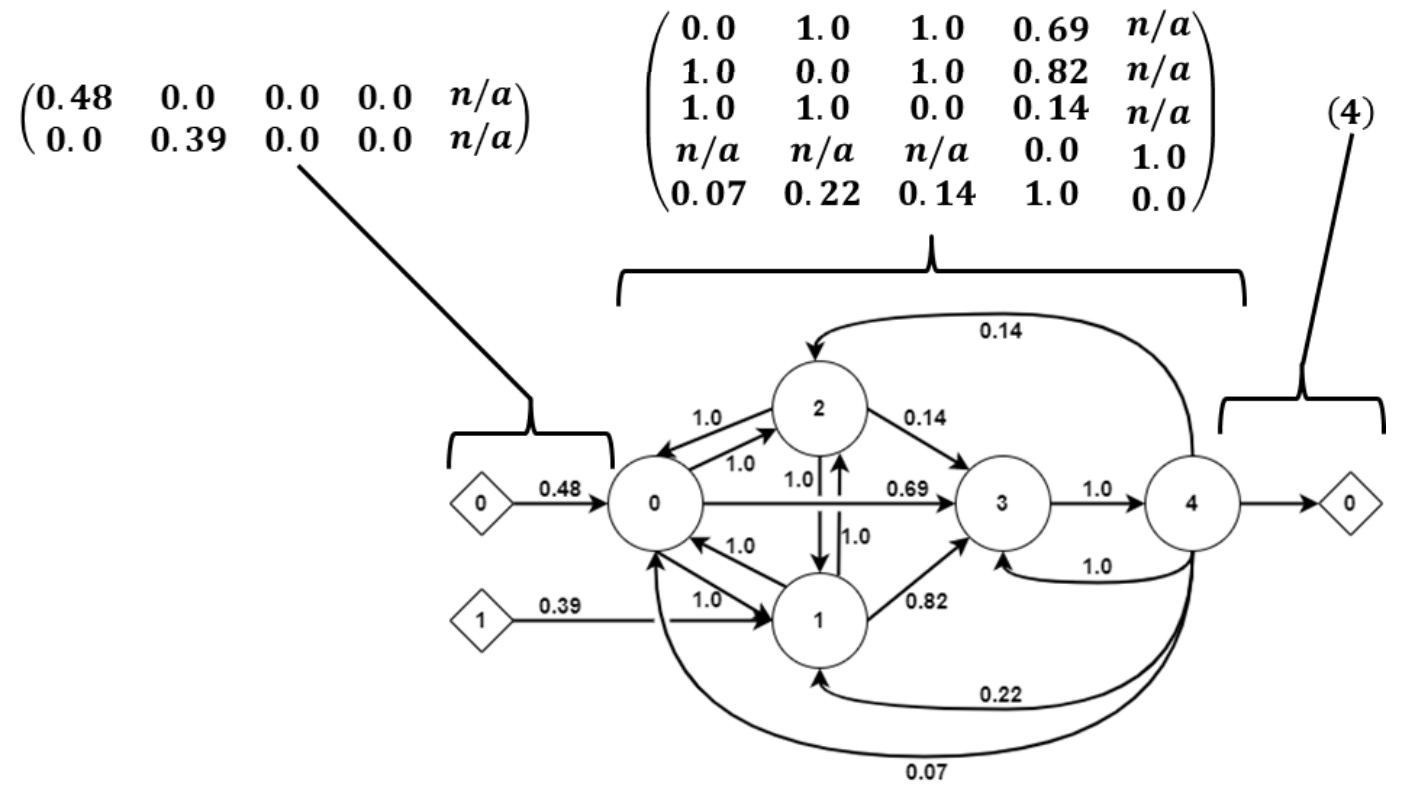}
    \caption{An Example of a two Motif Microcircuit, consisting of a CPG \& FBE with two input and output connection}
    \label{fig:Example_Microcircuit}
\end{figure}

To complete the description of a Microcircuit, it is necessary to add two more components: a second matrix containing the input connections, and an array of output connections. An example of this can be seen in figure~\ref{fig:Example_Microcircuit}, which completes the outline started in figure~\ref{fig:Matrix_Tiling_Example}.

\subsection{Behavioral Assessment}

Once a selection of prospective Microcircuits exists, it is necessary to assess their behaviour. This step is conceptually straightforward, as all that is required is to record the response of the various Microcircuits to a given stimulus or set of stimuli. Ideally, there would be a standard stimulus or set of stimuli which could provide a sufficiently comprehensive set of responses from the Microcircuit so as to consider its behaviour fully mapped, i.e., all possible forms of output from that Microcircuit have been solicited and recorded. However, the creation of such a test stimulus or set of stimuli has not yet been undertaken, hence preliminary experiments employed a less generic stimulus pattern, described later in this paper.

\subsection{Catalogue Population}

Regardless of the exact test stimuli used, the result is a set of Microcircuits, each with one or more output spike trains which illustrate their response to the stimuli. Using these, it is then necessary to determine whether a Microcircuit is suitable for inclusion in the component catalogue. To increase the coverage of Microcircuits, Novelty Search is used to produce a catalogue that encompasses a sufficiently large breadth of the behavioural space.

Novelty Search is an alternative to fitness function based evolutionary algorithms \cite{Bartz-2014}. Instead of assessing individuals based on how successful their behaviours are at meeting some explicitly defined criteria, they are instead compared on the basis of how different their behaviours are from one another. Individuals surpassing some thresholds of difference are then added to an archive, which forms the output of the search \cite{Lehman-2008}. Originally this archive was then searched to locate a final “best” individual, but in this use case, the archive constitutes the desired final product: a catalogue of Microcircuits with that display a diverse range of different behaviours. This approach also allows for a catalogue to be created without needing to characterize any of the desired behaviours or even without knowing what those behaviours might be.

\begin{equation} \label{novelty_equation}
    p(x) = \frac{1}{k}\sum\limits_{i=0}^k D_s^{x,i}
\end{equation}

\begin{equation} \label{archive_addition_equation}
    arc_{add}(x) = \begin{cases}
                        True \quad if \quad p(x) \geq p_{threshold} \\
                        else \quad false
                   \end{cases}
\end{equation}

\begin{equation} \label{threshold_equation}
    p_{threshold} = \begin{cases}
                        p_{threshold} * 0.95 \quad if \quad eval_{without arch} \geq A \\
                        p_{threshold} * 1.20 \quad if \quad added_{over C} \geq B \\
                        else \quad p_{threshold}
                   \end{cases}
\end{equation}

The equations describing the metric and threshold of the novelty search process are shown in equations~\ref{novelty_equation} \& ~\ref{archive_addition_equation}. The first is the novelty metric $p(x)$ also referred to as Sparseness, as it relates to how far removed an individual is from other individuals within the behavioural space, with the number of individuals being compared being described as the k-nearest neighbours. $D_s^{x,i}$ is the distance between the individual being assessed, x, and another individual i, and is task specific. As stated before, addition to the archive requires $p(x)$ to be above a threshold value, $p_{threshold}$. This threshold however is dynamic, being controlled by equation~\ref{threshold_equation}. If the search progresses for A generations without an addition to the archive, the threshold is lowered by 5\%; alternativly, if B individuals are added to the archive within a C generations, the threshold is raised by 20\%.

To employ this method, a means of computing the distance between the behaviours of two
Microcircuits is required. Fortunately, neuroscientists have need of something similar for comparing the recorded spike trains of biological neurons and have produced a wide range of methods that can be applied in this work, coincidentally referred to as Spike Distance Metrics \cite{Kreuz-2011}. Bivariate SPIKE-Distance has been chosen, a measure that has the advantages of taking a wide range of spike train metrics into consideration, while producing a single value. This value is between 0.0 and 1.0 and corresponds to the “distance” between the two spike trains, with a value of 0.0 only being returned if the two spike trains are identical \cite{Kreuz-2010}.

\begin{figure}[H]
    \centering
    \includegraphics[width=0.6\linewidth]{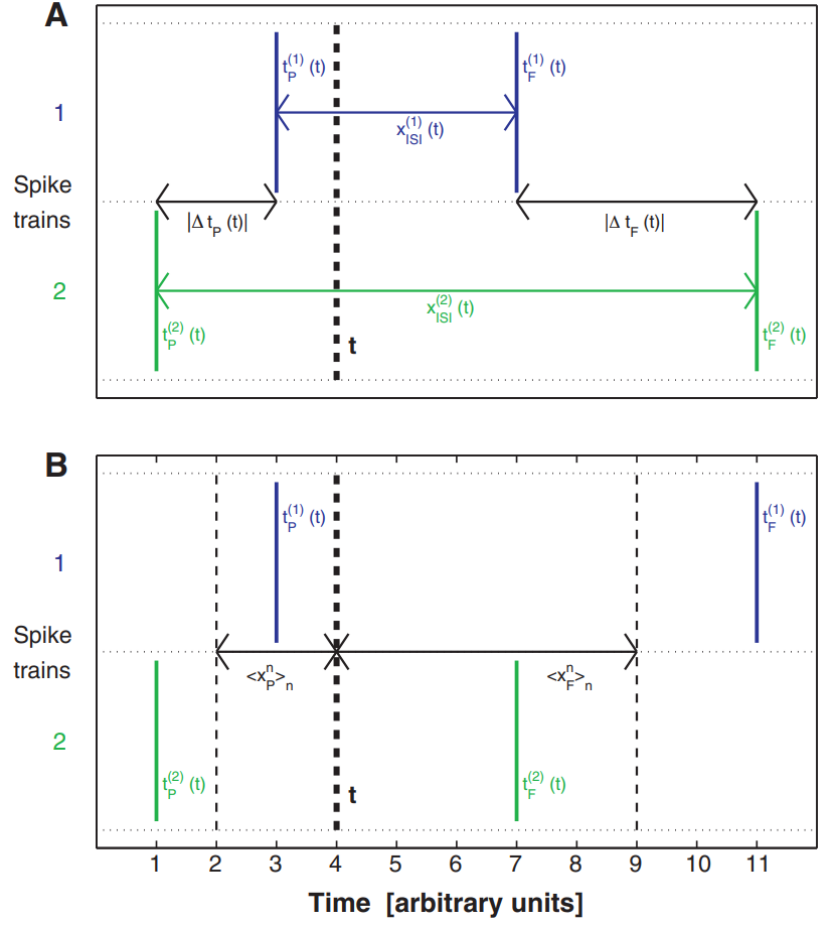}
    \caption{Illustration of the measurements taken at each time step for computing Bivariate SPIKE-Distance, as well as a number of the values calculated from those measurements}
    \label{fig:BSD_Plots}
\end{figure}

To compute the Bivariate SPIKE-Distance between two spike trains, the trains are first divided up into series of time steps, which must be smaller than the smallest inter-spike interval (ISI). At each of these time steps, three measurements are taken from the two spike trains (as shown in figure~\ref{fig:BSD_Plots}A): the time at which the last spike occurred $t_p^{(n)}(t)$; the time at which the next spike will occur $t_f^{(n)}(t)$; and the inter-spike interval of those two spikes $X_{ISI}^{(n)}(t)$. From these values, five parameters at then calculated: the absolute instantaneous difference between the previous spike times $|\Delta t_p(t)|$ (equation ~\ref{Absolute instantaneous difference of previous spike times}); the absolute instantaneous difference between the following spike times $|\Delta t_f(t)|$ (equation~\ref{Absolute instantaneous difference of following spike times}); the average previous spike interval $\langle X_p^{(n)}(t) \rangle_n$ (equation~\ref{Average previous spike interval}); the average following spike interval $\langle X_f^{(n)}(t) \rangle_n$ (equation ~\ref{Average following spike interval}); and the average interspike interval $\langle X_{ISI}^{(n)}(t) \rangle_n$ (equation~\ref{Average interspike interval}). These are then in turn used to compute the Instantaneous Spike Difference $S(t)$ using equation~\ref{Instantaneous Spike Difference equation}. By computing this instantaneous difference for each time step, and then integrating over the array of resulting values, a final Bivariate SPIKE-Distance value for the two spike trains is acquired.

\begin{equation} \label{Absolute instantaneous difference of previous spike times}
    |\Delta t_p(t)| =  |t_p^{(1)}(t) - t_p^{(2)}(t)|
\end{equation}

\begin{equation} \label{Absolute instantaneous difference of following spike times}
    |\Delta t_f(t)| =  |t_f^{(1)}(t) - t_f^{(2)}(t)|
\end{equation}

\begin{equation} \label{Average previous spike interval}
    \langle X_p^{(n)}(t) \rangle_n = \frac{1}{n}\sum\limits_{n=1}^Nt - t_p^{(n)}(t)
\end{equation}

\begin{equation} \label{Average following spike interval}
    \langle X_f^{(n)}(t) \rangle_n = \frac{1}{n}\sum\limits_{n=1}^Nt_f^{(n)}(t) - t
\end{equation}

\begin{equation} \label{Average interspike interval}
    \langle X_{ISI}^{(n)}(t) \rangle_n = \frac{1}{n}\sum\limits_{n=1}^NX_{ISI}^{(n)}(t)
\end{equation}

\begin{equation} \label{Instantaneous Spike Difference equation}
    S(t) = \frac{|\Delta t_p(t)|\langle X_f^{(n)}(t) \rangle_n + |\Delta t_f(t)|\langle X_p^{(n)}(t) \rangle_n}{\langle X_{ISI}^{(n)}(t) \rangle_n^2}
\end{equation}

By applying this measure to each unique paring of Microcircuits, a distance matrix can be
generated, which provides the values needed to compute the sparseness values required by
the novelty search algorithm.

\begin{figure}[H]
    \centering
    \includegraphics[width=0.9\linewidth]{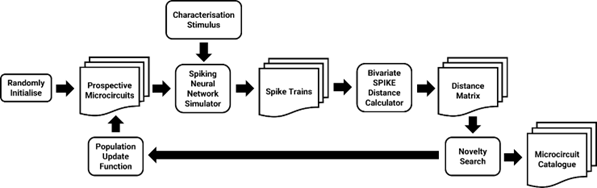}
    \caption{An illustration of the complete Microcircuit catalogue generation methodology}
    \label{fig:Generator_Methology}
\end{figure}

All that remains is to iterate the population of prospective Microcircuits in the standard fashion (using crossover \& mutation as part of selection), before repeating the process to produce a suitable catalogue, as illustrated in figure~\ref{fig:Generator_Methology}.

\section{Proof-of-Concept Experiment}
To establish if this methodology is sufficient to produce a selection of useful Microcircuits (that is to say Microcircuits whose behaviours provide some functionality that is a desirable component of the overall functioning of a larger neuromorphic system), it is necessary to put it to the test. For this purpose an example input stimulus must be chosen, one which would present a range of possible input patterns that can be mapped, at least conceptually, to some hypothetical application. An emulation of an 8-bit bus was selected for this purpose, with the intended output therefore being Microcircuits that displayed clear and consistent responses to different bit patterns of the data.

\subsection{Stimulus}
To provide an input stimulus that is meaningful in both quantity and quality, this experiment used a sample of text. This sample consisted of 2030 characters, including spaces, complete with punctuation. Each character of the text sample was translated to a single byte, following the UTF-8 encoding scheme, with these binary values then used to produce a set of eight input spike trains. A simple encoding scheme was used, where a binary 1 produced a 25ms long burst of spikes, at a rate of 1 spike per ms, on the associated input channel (25ms was selected as it was the same as $\tau$ of the example spiking neuron being used).

\begin{figure}[H]
    \centering
    \includegraphics[width=0.65\linewidth]{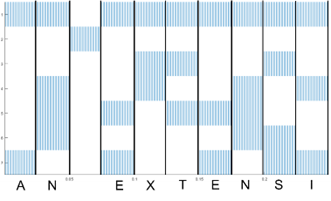}
    \caption{A section of the input stimulus. Note that there are only 7 input spike trains shown, as UTF-8 does not use the Least Significant Bit (LSB) for all characters.}
    \label{fig:Input_Stimulus}
\end{figure}

It should be noted that, while in the context of this experiment it is specific characters that are being considered, any Microcircuits that display the desired kind of behaviours will be doing so because they have latched onto the pattern of input spike trains those characters have been mapped to. Those patterns could correspond to other stimuli of interest, for example: control codes on a microprocessor’s instruction bus. Furthermore, validating the ability of this approach to reliably produce Microcircuits that latch onto patterns in a useful fashion would indicate its broad applicability, as not only could any conceivable input of interest be encoded thus, but the outputs of the Microcircuits themselves are spike train patterns, meaning that there is nothing to prevent them being assembled sequentially.    

\subsection{Parameters}
The population of Microcircuits consisted of 100 individuals and was iterated 50 times. All Microcircuits had eight inputs and one output, with the starting pool of Microcircuits all having two Motifs. Input and internal connections had a 25\% chance of existing (tuned to allow for suitably sparse connectivity) and a weight value between -1.0 and 1.0 (with negative weights corresponding to inhibitory synapses \& postive corrisponding to exitatory ones).

Each iteration of the population produced new Microcircuits by combining two Microcircuits from the previous iteration: one with a high average Bivariate SPIKE-Distance and one with a low average. This avoided the possibility of self-crossover, while also aiding diversity. Crossover ratio was 4:6 in favor of the low SPIKE-Distance individual. Following crossover, mutation was applied with the following probabilities: 35\% chance to add a new Motif; 60\% chance to replace a Motif; 60\% chance to alter the weight of an input or internal connection; and a 60\% chance of changing an output connection. These values were deliberately set quite high so as to fuel diversity within the population.

To be added to the catalogue, a Microcircuit needed to possess an average Bivariate SPIKE-Distance of at least 0.5, though this value was dynamic, as suggested in (Lehman \& Stanley, 2008). If 10 iterations of Microcircuits passed without an addition being made to the catalogue, the threshold was reduced by 5\%. Alternatively, if 10 individuals were added to the catalogue in the space of one iteration, the threshold was raised by 20\%.

\subsection{Results}
From a total pool of 5000 Microcircuits created over the course of the experiment, 50 were deemed to display sufficiently novel behaviour to warrant transfer to the catalogue, with average Bivariate SPIKE-Distance values of between 0.44 and 0.63. As is to be expected from the increase in complexity that occurs in novelty searches, the number of motifs, and as such neurons, within the Microcircuits increased steadily, with the largest instance coming from generation 45: possessing 61 neurons across 20 motifs.

\begin{figure}[H]
    \centering
    \includegraphics[width=0.95\linewidth]{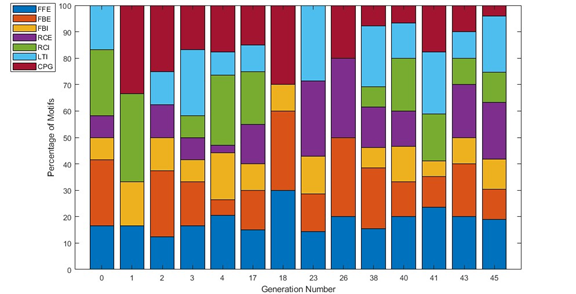}
    \caption{Percentage breakdown of the motif makeup of Microcircuits in the catalogue, divided based on the generation the Microcircuit originates from.}
    \label{fig:Motif_Makeup}
\end{figure}

The breakdown of which motifs are present in the Microcircuits added to the catalogue is shown in figure~\ref{fig:Motif_Makeup}. Given that there isn't a preponderance of one kind of Motif over another, and that over variation across the generations shows no clear trend, the data suggests there is no inclination towards a specific ratio of motifs for novel microcircuits, nor that the presence or absence of a given motif is of any specific importance.

To determine if any of the Microcircuits in the catalogue demonstrated clearly useful behaviour, a correlation was determined between the output spike trains of the Microcircuits and the character the input spike trains were presenting. To do this, the average number of output spikes across each of the input characters was calculated. If this value was zero, the Microcircuit was not responding to a given character; values less than one would indicate that a character was only responded to sometimes (a weak correlation); while any value greater than one would indicate the Microcircuit spiked more than once on each instance of a character (a strong correlation)

\begin{figure}[H]
    \centering
    \begin{subfigure}[t]{0.4\columnwidth}
        \includegraphics[width=0.95\linewidth]{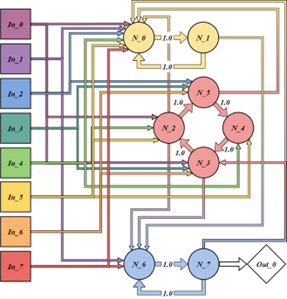} 
        \caption{}
        \label{fig:Microcircuit_392}
    \end{subfigure}
    \begin{subfigure}[t]{0.5\columnwidth}
        \includegraphics[width=0.95\linewidth]{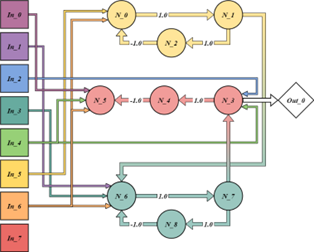} 
        \caption{}
        \label{fig:Microcircuit_466}
    \end{subfigure}
    \caption{Illustrations of (a) Microcircuit 392 \& (b) Microcircuit 466}
    \label{fig:Microcircuits_392_466}
\end{figure}

Across all 50 Microcircuits in the catalogue, 8 Microcircuits exhibited strong correlations for all characters, which on closer investigation corresponded to different forms of consistent spiking. Of the others: 20 Microcircuits displayed weak correlations to one or more characters; 15 displayed a mixture of strong and weak correlations, while the remaining 7 displayed strong correlations with specific subsets of characters. Of these 7, two are of particular note: Microcircuits 392 \& 466 (figure~\ref{fig:Microcircuits_392_466}).

Microcircuit 392, produced in the fourth generation with 8 neurons across 3 motifs (FBE, RCI \& FBE), displayed a strong correlation (average of 1.20 spikes) with the dashes in the sample text.

Microcircuit 466, produced in the fifth generation with 9 neurons across 3 motifs (FBI, LTI \& FBI), displayed an even stronger correlation with all four non-letter characters in the sample text (Space: 4.31; Comma: 4.50; Dash: 3.80; Stop: 4.33).

\section{Expanded Search Experiment}
After the success of the proof-of-concept experiment, a few modifications were
made to the algorithm and the first attempt to generate a meaningful catalogue of useful
Microcircuits was carried out. Aside from minor parameters tweaks, three modifications
were made: the switch to an adjusted set of Motifs; the addition of a pruning algorithm to the population iteration process; and the replacement of the original test stimulus.

\subsection{Adjusted Motif Selection}
The selection of Motifs utilised for the initial experiment was a simplified subset of the Motifs reported by \cite{Luo-2016}. Therefore, in the interests of expanding the range of possible behaviours that can be obtained, the next step is to employ all ten of Luo's identified Motifs: FeedForward Excitation (FFE), FeedForward Inhibition (FFI); FeedBack Excitation (FBE); FeedBack Inhibition (FBI); FeedForward Recurrent Excitation (FFRE); FeedForward Recurrent Inhibition (FFRI); FeedBack Recurrent Excitation (FBRE); FeedBack Recurrent Inhibition (FBRI); FeedForward Lateral Inhibition (FFLI); and FeedBack Lateral Inhibition (FBLI).

\begin{figure}[H]
    \centering
    \includegraphics[width=0.5\columnwidth]{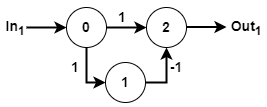}
    \caption{The FeedForward Inhibition (FFI) Motif.}
    \label{fig:FFI}
\end{figure}

\begin{figure}[H]
    \centering
    \begin{subfigure}[t]{0.45\columnwidth}
        \includegraphics[width=1\linewidth]{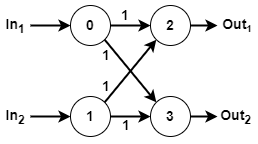} 
        \caption{}
        \label{fig:FFRE}
    \end{subfigure}
    \begin{subfigure}[t]{0.45\columnwidth}
        \includegraphics[width=1\linewidth]{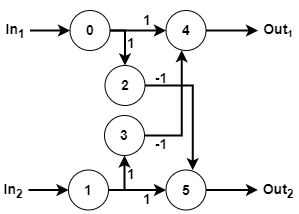} 
        \caption{}
        \label{fig:FFRI}
    \end{subfigure}
    \caption{The (a) FeedForward Recurrent Excitation (FFRE) \& (b) FeedForward Recurrent Inhibition (FFRI) Motifs}
    \label{fig:FFRE_FFRI}
\end{figure}

\begin{figure}[H]
    \centering
    \begin{subfigure}[t]{0.45\columnwidth}
        \includegraphics[width=1\linewidth]{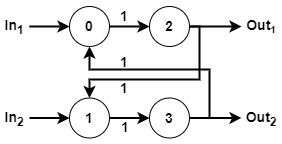} 
        \caption{}
        \label{fig:FBRE}
    \end{subfigure}
    \begin{subfigure}[t]{0.45\columnwidth}
        \includegraphics[width=1\linewidth]{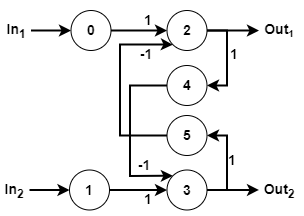} 
        \caption{}
        \label{fig:FBRI}
    \end{subfigure}
    \caption{The (a) FeedBack Recurrent Excitation (FBRE) \& (b) FeedBack Recurrent Inhibition (FBRI) Motifs}
    \label{fig:FBRE_FBRI}
\end{figure}

\begin{figure}[H]
    \centering
    \begin{subfigure}[t]{0.45\columnwidth}
        \includegraphics[width=1\linewidth]{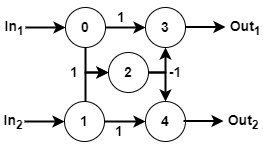} 
        \caption{}
        \label{fig:FFLI}
    \end{subfigure}
    \begin{subfigure}[t]{0.45\columnwidth}
        \includegraphics[width=1\linewidth]{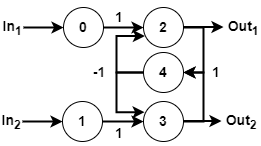} 
        \caption{}
        \label{fig:FBLI}
    \end{subfigure}
    \caption{The (a) FeedForward Lateral Inhibition (FFLI) \& (b) FeedBack Lateral Inhibition (FFLI) Motifs}
    \label{fig:FFLI_FBLI}
\end{figure}

\begin{table*}[bth]
    \centering
    \begin{tabular}{|c|c|c|c|}
         \noalign{\hrule height 1pt}
        Motif & Connection Matrix & Input Neurons & Output Neurons \\
         \noalign{\hrule height 1.5pt}
        \shortstack{FeedForward Inhibition\\ (FFI)} & \centering \begin{adjustbox}{valign=m} $ \begin{bmatrix}
                            0 & 1 & 1  \\
                            0 & 0 & -1 \\
                            0 & 0 & 0
                        \end{bmatrix} $ \end{adjustbox} & 0 & 2 \\
         \noalign{\hrule height 1pt}
        \shortstack{FeedForward Recurrent\\ Excitation (FFRE)} &  \centering \begin{adjustbox}{valign=m}  $ \begin{bmatrix}
                            0 & 0 & 1 & 1 \\
                            0 & 0 & 1 & 1 \\
                            0 & 0 & 0 & 0 \\
                            0 & 0 & 0 & 0
                        \end{bmatrix} $ \end{adjustbox} & 0,1 & 2,3 \\
         \noalign{\hrule height 1pt}
       \shortstack{FeedForward Recurrent\\ Inhibition (FFRI)} &  \centering \begin{adjustbox}{valign=m}  $ \begin{bmatrix}
                            0 & 0 & 1 & 0 & 1 & 0  \\
                            0 & 0 & 0 & 1 & 0 & 1  \\
                            0 & 0 & 0 & 0 & 0 & -1 \\
                            0 & 0 & 0 & 0 & -1 & 0 \\
                            0 & 0 & 0 & 0 & 0 & 0  \\
                            0 & 0 & 0 & 0 & 0 & 0  
                         \end{bmatrix} $ \end{adjustbox} & 0,1 & 4,5 \\
         \noalign{\hrule height 1pt}
        \shortstack{FeedBack Recurrent\\ Excitation (FBRE)} & \centering \begin{adjustbox}{valign=m}  $ \begin{bmatrix}
                            0 & 0 & 1 & 0 \\
                            0 & 0 & 0 & 1 \\
                            0 & 1 & 0 & 0 \\
                            1 & 0 & 0 & 0
                        \end{bmatrix} $ \end{adjustbox} & 0,1 & 2,3 \\
        \noalign{\hrule height 1pt}
        \shortstack{FeedBack Recurrent\\ Inhibition (FBRI)} & \centering \begin{adjustbox}{valign=m}  $ \begin{bmatrix}
                            0 & 0 & 1 & 0 & 0 & 0  \\
                            0 & 0 & 0 & 1 & 0 & 0  \\
                            0 & 0 & 0 & 0 & 1 & 0  \\
                            0 & 0 & 0 & 0 & 0 & 1  \\
                            0 & 0 & 0 & -1 & 0 & 0  \\
                            0 & 0 & -1 & 0 & 0 & 0  
                        \end{bmatrix} $ \end{adjustbox} & 0,1 & 2,3 \\
        \noalign{\hrule height 1pt}
        \shortstack{FeedForward Lateral\\ Inhibition (FFLI)} &  \centering \begin{adjustbox}{valign=m}  $ \begin{bmatrix}
                            0 & 0 & 1 & 1 & 0   \\
                            0 & 0 & 1 & 0 & 1   \\
                            0 & 0 & 0 & -1 & -1 \\
                            0 & 0 & 0 & 0 & 0   \\
                            0 & 0 & 0 & 0 & 0   
                         \end{bmatrix} $ \end{adjustbox} & 0,1 & 3,4 \\ 
        \noalign{\hrule height 1pt}
        \shortstack{FeedBack Lateral\\ Inhibition (FBLI)} &  \centering \begin{adjustbox}{valign=m}  $ \begin{bmatrix}
                            0 & 0 & 1 & 0 & 0   \\
                            0 & 0 & 0 & 1 & 0   \\
                            0 & 0 & 0 & 0 & 1 \\
                            0 & 0 & 0 & 0 & 1   \\
                            0 & 0 & -1 & -1 & 0  
                         \end{bmatrix} $ \end{adjustbox} & 0,1 & 2,3 \\
        \noalign{\hrule height 1pt}
    \end{tabular}
    \caption{\label{tab:table-name-second} Connection Matrices and valid Input/Output neurons for the new Motifs}
    \label{table:Motif_Matrices_Two}
\end{table*}

\subsection{Pruning Algorithm}
In the preliminary experiment, the mutations that could be applied by the iteration function were the addition or alteration of Motifs; the alteration of Input connections; or the alteration of the Output connections. This, coupled with the drive of Novelty Search functions towards more complex structures to produce new behaviours, resulted in a steady growth in the size of new Microcircuits. However, the Microcircuits that displayed the most useful novel behaviours emerged early on and as such were smaller in size. This, coupled with the fact that the biological examples articulated previously are also smaller in size, suggests a need to constrain the growth of the Microcircuits. Thus, a pruning function (Algorithm~\ref{alg:pruning}) was introduced.

The first step of the pruning process is the computation of a Complexity Value for the
population currently being iterated. This is simply the mean number of neurons \&
connections within the Microcircuits in that population. This value is then compared to one of two thresholds, depending on which mode the iterator is in. If the iterator is currently functioning normally (i.e. mutations add Motifs), and the complexity if above the Start Pruning Threshold, the iterator switches to the pruning mode (i.e. mutations remove Motifs). Conversely, if the iterator is currently in the pruning mode, and the complexity has dropped below the End Pruning Threshold, the iterator returns to the normal mode.

\begin{algorithm}
    \caption{operation of the pruning algorithm within the population iterator.}\label{alg:pruning}
    \begin{algorithmic}
        \ForAll{$Microcircuits \in Population$}
            \State $Complexity\gets number\ of\ connections + number\ of\ neurons$
        \EndFor
        \State $Calculate \ mean \ complexity$
        
        \If{$Iterator\ is\ not\ pruning$}
            \If{$mean\ complexity \geq start\ threshold$}
                \State $Iterator \gets Pruning\ mode$
            \Else
                \State $Iterator \gets Standard\ mode$
            \EndIf
        \EndIf
    
        \If{$Iterator\ is\ pruning$}
            \If{$mean\ complexity \leq end\ threshold$}
                \State $Iterator \gets Standard\ mode$
            \Else
                \State $Iterator \gets Pruning\ mode$
            \EndIf
        \EndIf
    \end{algorithmic}
\end{algorithm}

\subsection{Stimulus}
The test stimulus employed the same basic methodology as its predecessor: the use of a text
string to generate a set of 8 spike trains that encode a series of spatial patterns. However, the text string used was longer, consisting of 2969 characters produced using extended Lorum Ipsum, resulting in a total of 54 unique characters. This was done to provide a wider range of input stimulus patterns. 

\subsection{Parameters}
The population of Microcircuits consisted of 100 individuals and was iterated 1000 times. All Microcircuits had eight inputs and one output, with the starting pool of Microcircuits all having two Motifs. Input and internal connections had a 25\% chance of existing (tuned to allow for suitably sparse connectivity) and a weight value between -1.0 and 1.0.

Crossover occurred as outlined in the previous experiment, with a 4:6 crossover ratio followed by mutation with the following probabilities: 35\% chance to add a new Motif or 40\% chance to remove a new Motif (depending on mode); 75\% chance to replace a Motif; 60\% chance to alter the weight of an input or internal connection; and a 60\% chance of changing an output connection. A population complexity over 60 switched the iterator from normal to pruning mutation, while a complexity below 40 switched it back.

The initial threshold for inclusion in the novelty search archive was 0.5, with this value
decreasing by 5\% after 20 generations without an addition and increasing by 20\% if 10
individuals were added in a single generation. The number of neighbours was set to 100.

\subsection{Results}
From the pool of 100,000 Microcircuits created over the course of the experiment, 993 were
added to the archive: with average Bivariate SPIKE-Distance values of between 0.54 and 0.58. A sampling of Microcircuits taken from the archive reveal an average of 11 neurons per Microcircuit across 2 to 5 Motifs. A clear indication that the pruning function worked as intended.

Analysis of the output spike trains of the Microcircuits revealed a significant issue however: the majority of the 993 Microcircuits demonstrated strong spiking correlations with the majority of characters (over 79\%), which closer investigation confirmed to be due to a range of consistent spiking behaviours. A sample of Microcircuits are shown in table~\ref{table:Sample_Microcircuits} to illustrate this issue.

\begin{table*}[bth]
    \centering
    \begin{tabular}{|C{1.2cm}|C{2.2cm}|C{1.9cm}|C{1.2cm}|C{5cm}|C{2.2cm}|}
         \noalign{\hrule height 1pt}
         ID & Generation & Neurons & Motifs & Spikes On & Percentage \\
         \noalign{\hrule height 1pt}
         9819 & 98 & 17 & 5 & SPACE; !; COMMA; STOP; ?; A; B; C; D; E; F; G; H; I; L; M; N; O; P; Q; R; S; T; U; V; Z; a; b; c; d; e; f; g; h; i; l; m; n; o; p; q; r; s; t; u; v; x; á; é; ó & 92.59\% \\
         \noalign{\hrule height 1pt}
         12003 & 120 & 15 & 4 & SPACE; !; COMMA; STOP; ?; A; B; C; D; E; F; G; H; I; L; M; N; O; P; Q; R; S; T; U; V; Z; a; b; c; d; e; f; g; h; i; l; m; n; o; p; q; r; s; t; u; v; x; Á; É; á; é; ó & 96.30\% \\
         \noalign{\hrule height 1pt}
         28862 & 288 & 10 & 4 & SPACE; !; COMMA; ?; A; B; C; D; E; F; H; I; L; M; N; O; P; Q; R; S; T; U; V; Z; a; b; c; d; e; f; g; h; i; l; m; n; o; p; q; r; s; t; u; v; x; É; á; é; ó & 90.74\% \\
         \noalign{\hrule height 1pt}
         50507 & 505 & 6 & 2 & SPACE; COMMA; ?; A; C; D; E; H; I; L; M; N; O; P; Q; S; T; U; V; a; b; c; d; e; f; g; h; i; l; m; n; o; p; q; r; s; t; u; v; x; á; é; ó & 79.63\% \\
         \noalign{\hrule height 1pt}
         76103 & 761 & 9 & 2 & SPACE; !; COMMA; STOP; ?; A; C; D; E; F; G; H; I; L; M; N; P; Q; R; S; T; U; V; a; b; c; d; e; f; g; h; i; l; m; n; o; p; q; r; s; t; u; v; x; á; é; ó & 87.04\% \\
         \noalign{\hrule height 1pt}
    \end{tabular}
    \caption{\label{tab:table-name} Details of sample Microcircuits}
    \label{table:Sample_Microcircuits}
\end{table*}

\begin{figure}[H]
    \centering
    \includegraphics[width=\columnwidth]{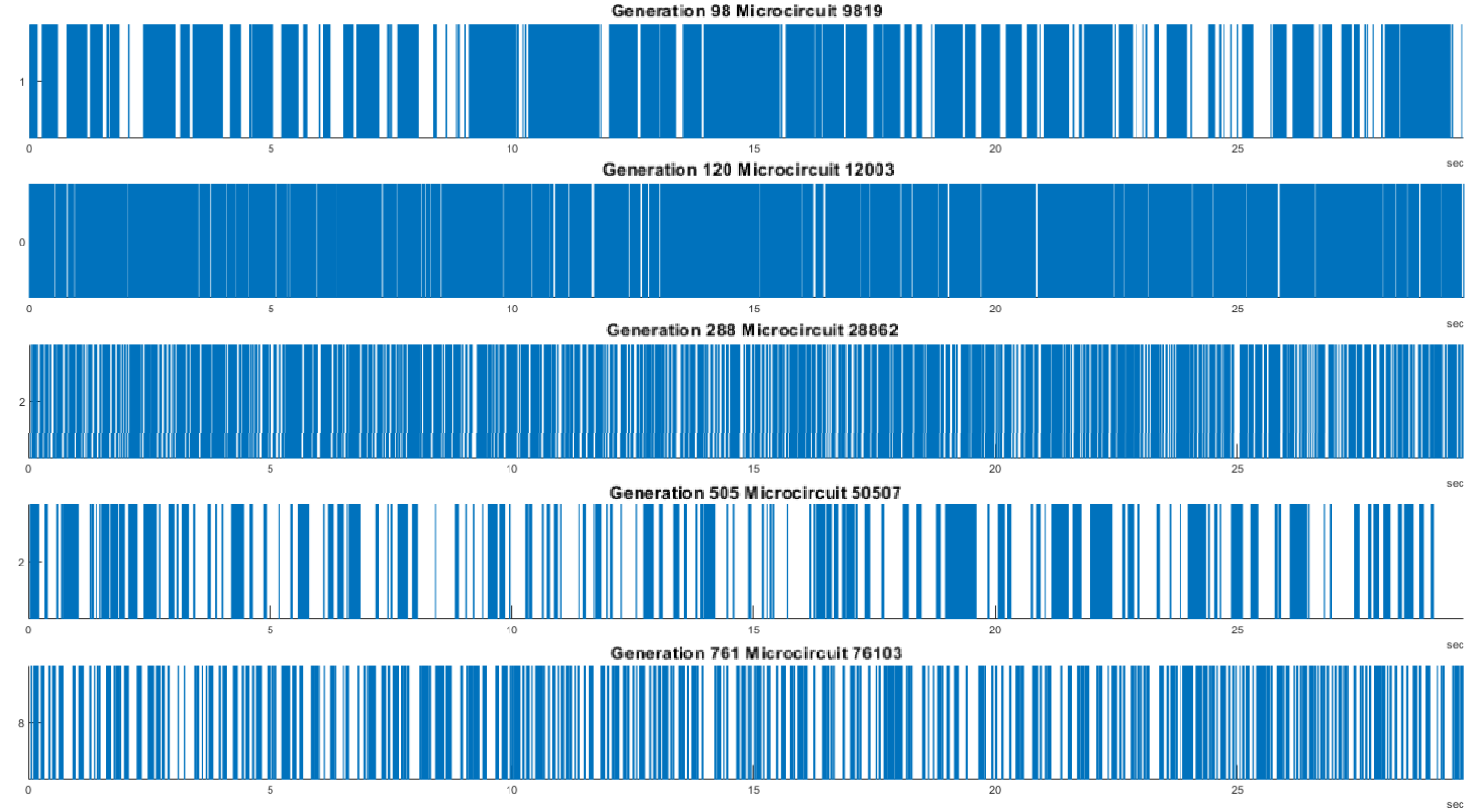}
    \caption{Output spike trains of the sample Microcircuits.}
    \label{fig:output_samples}
\end{figure}

\begin{figure}[H]
    \centering
    \includegraphics[width=0.95\columnwidth]{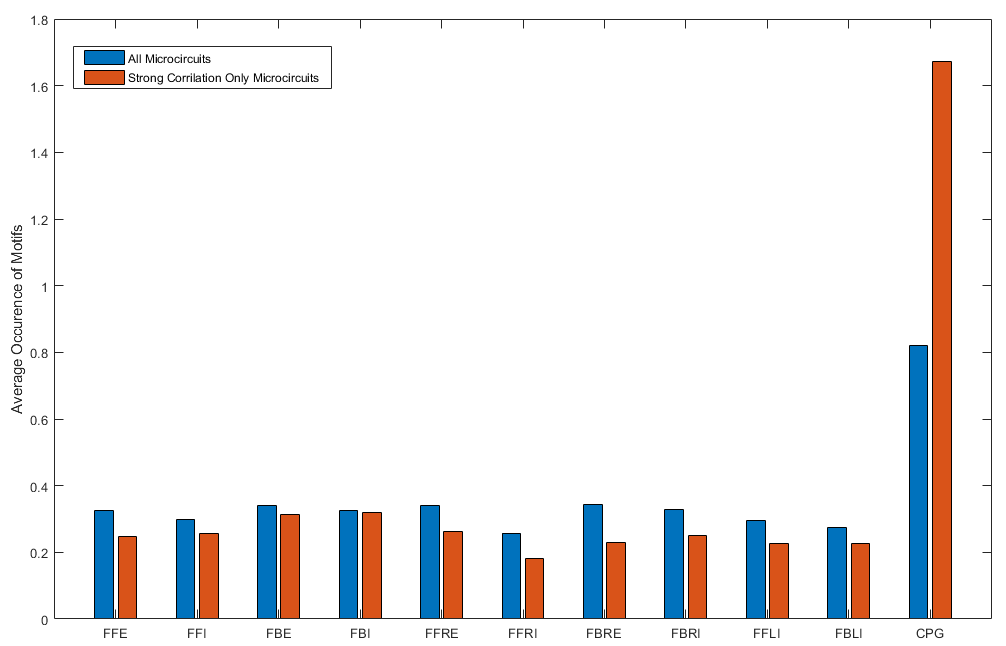}
    \caption{Breakdown of the average occurrence of Motifs within the Microcircuit archive. Blue is the average across all Microcircuits; Red across only those Microcircuits demonstrating the oscillatory behaviour.}
    \label{fig:motif_breakdown_two}
\end{figure}

To try an ascertain the cause of this issue, an analysis of the Motifs present within the
Microcircuits was performed, both the complete archive and those Microcircuits which exhibited the undesirable oscillatory behaviour. The results of this this analysis are shown in figure~\ref{fig:motif_breakdown_two}

This analysis leads to a clear conclusion: Central Pattern Generators are overrepresented within the population, a conjecture born out by the fact that all five of the sample Microcircuits in table~\ref{table:Sample_Microcircuits} contain a CPG, while investigation of the wider archive revealed Microcircuits with as many as three CPGs.

However, while the immediate remedy is simple; the removal of the CPGs from the pool of possible Motifs, a more crucial matter has been highlighted. These oscillatory behaviours are obviously \textit{novel}, otherwise they would not be present in the archive, but they are not \textit{useful}. That is to say, there is no clear application for their behaviours within a system, especially when contrasted with the Microcircuits produced in the experiment outlined in section 4. Furthermore, given the significant disparity between the utility of these Microcircuits and those produced during the proof of concept experiment in section 4 it suggests that one of the changes articulated previously in this section is responsible.   

\section{Evaluation and Analysis of the Generator Methodology}
To isolate the specific change responsible for the issue, experiments were to be carried out where the various modifications to the generator were reverted. As it represented one of the more significant differences, the first reversion was to the stimulus used for the original proof-of-concept experiment from section 4.

\subsection{Original Stimulus Experiment}
The population of Microcircuits consisted of 100 individuals and was iterated 100 times. All Microcircuits had eight inputs and one output, with the starting pool of Microcircuits all having two Motifs. Input and internal connections had a 25\% chance of existing (tuned to allow for suitably sparse connectivity) and a weight value between -1.0 and 1.0.

Crossover occurred as outlined in the previous experiments, with a ratio of 4:6. Mutation then occurred with the following probabilities: 35\% chance to add a new Motif or 40\% chance to remove a new Motif (depending on mode); 75\% chance to replace a Motif; 60\% chance to alter the weight of an input or internal connection; and a 60\% chance of changing an output connection. A population complexity over 60 switched the iterator from normal to pruning mutation, while a complexity below 40 switched it back. The initial threshold for inclusion in the novelty search archive was 0.5, with this value decreasing by 5\% after 20 generations without an addition and increasing by 20\% if 10
individuals were added in a single generation. The number of neighbours was set to 25.

After execution, the returned archive of Microcircuits contained a number of individuals that responded with various strengths to different combinations of the Comma, Stop, Dash and Space characters: behaviours that were equally prevalent in the original proof of concept experiment from section 4. This clearly indicates two things: that the new stimulus added in section 5 is in some fashion deficient; but also that some character of the stimulus employed is correlated with Microcircuits produced.     

\subsection{Stimuli Analysis}
As a baseline, a hypothetical catalogue of Microcircuits would contain at least one Microcircuit for each distinct pattern of input, which in the case of the two stimulus sets that have previously been employed, corresponds to each unique character.  This of course does not exhaustively cover the possible novel Microcircuits for a given stimulus, such as those that responded to groups of input patterns (as with the results from section 4), but this "minimum viable catalogue" definition will suffice for this analysis. It is possible to construct an optimal response spike train for each of these hypothetical ideal Microcircuits, which would take the form of a set of bursts of spikes correlated with the instances of the respective input pattern in the stimulus.

\begin{figure}[H]
    \centering
    \includegraphics[width=0.95\columnwidth]{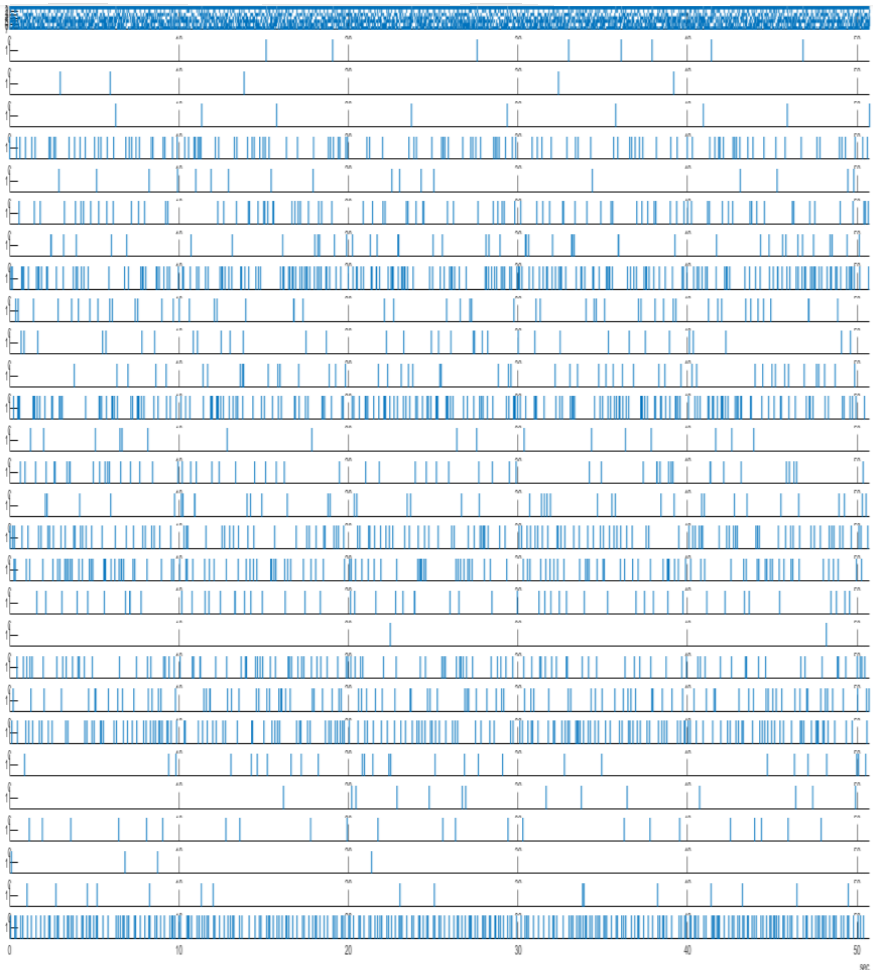}
    \caption{The complete stimulus spike trains (top) and derived Ideal Response Spike Trains for the characterisation stimulus used in the original proof-of-concept experiment.}
    \label{fig:ideal_response_spike_trains}
\end{figure}

The use of such Ideal Response Spike Trains, illustrated in figure~\ref{fig:ideal_response_spike_trains}, comes from the fact that they can be compared to one another using the Bivariate SPIKE-Distance measure, exactly as if they were the outputs of a population of Microcircuits being assessed. If the difference between two Ideal Response Spike Trains is significantly lower than the sparseness threshold (a.k.a << 0.5), that indicates that two real Microcircuits whose behaviours are close to those Hypothetical Microcircuits would likely not be considered novel enough to be promoted to the archive. In other words, if the Ideal Response Spike Trains of a given stimulus are not “sufficiently different” from one another, a run of the generator that uses that stimulus will not produce useful Microcircuits.

\begin{figure}[H]
    \centering
    \begin{subfigure}[t]{0.49\columnwidth}
        \includegraphics[width=1\linewidth]{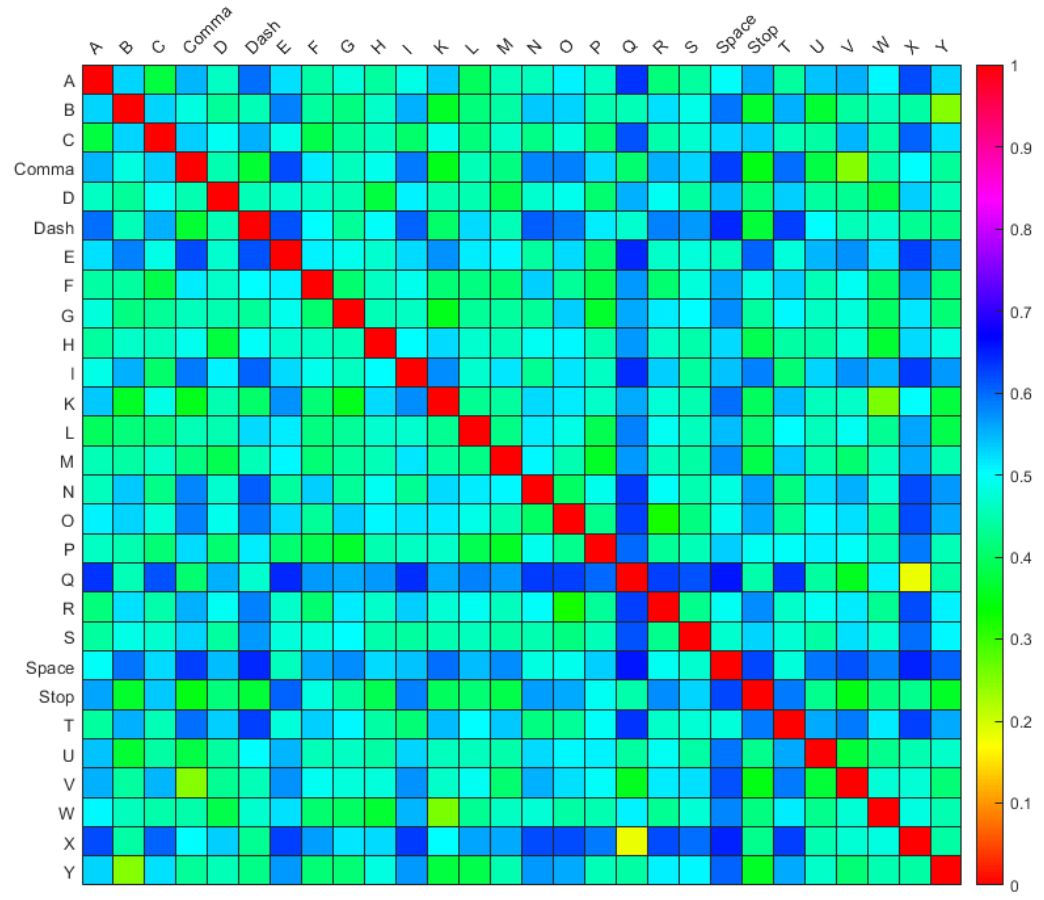} 
        \caption{}
        \label{fig:heatmap_one_full}
    \end{subfigure}
    \begin{subfigure}[t]{0.49\columnwidth}
        \includegraphics[width=1\linewidth]{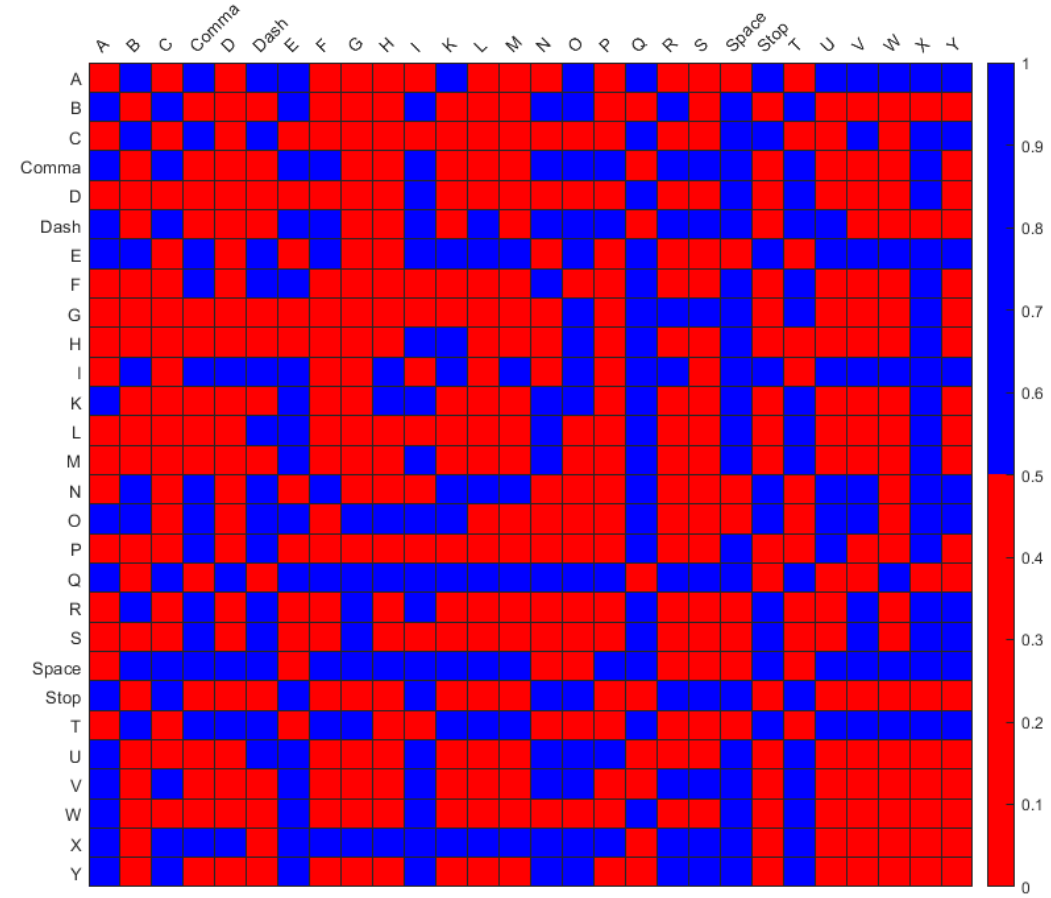} 
        \caption{}
        \label{fig:heatmap_one_thresholded}
    \end{subfigure}
    \caption{Heatmaps of the Bivariate SPIKE-Distances between the Ideal Response Spike Trains of the 28 unique characters of the first stimulus input. On Heatmap b) Red values are below the threshold (0.5), blue values are above}
    \label{fig:heatmaps_one}
\end{figure}

The two heatmaps in figure~\ref{fig:heatmaps_one} illustrate this ideal response cross-comparison for the original stimulus. A number of the input patterns have a reasonable degree of separability, with Comma \& Dash having 14 over threshold distances and Space having 20. Interestingly, Stop has only 9, which might suggest a reason why Microcircuits that recognised groups of punctuation characters were prevalent.    

By contrast, a number of the input patterns would be hard to generate Microcircuits for (e.g. ‘G’, 7; ‘H’, 4; \& ‘P’, 7) due to the low number of over threshold distances. It is interesting to note also the handful of characters with a high number of over threshold distances that did not result in corresponding Microcircuits, such as 'T' with 16; 'Q' with 20; \& 'X' with 19. In the latter two cases the infrequency of the characters is a possible cause, but given the prevalence of 'T' (174 instances) this is clearly not the whole story.  

\begin{figure}[H]
    \centering
    \begin{subfigure}[t]{0.49\columnwidth}
        \includegraphics[width=1\linewidth]{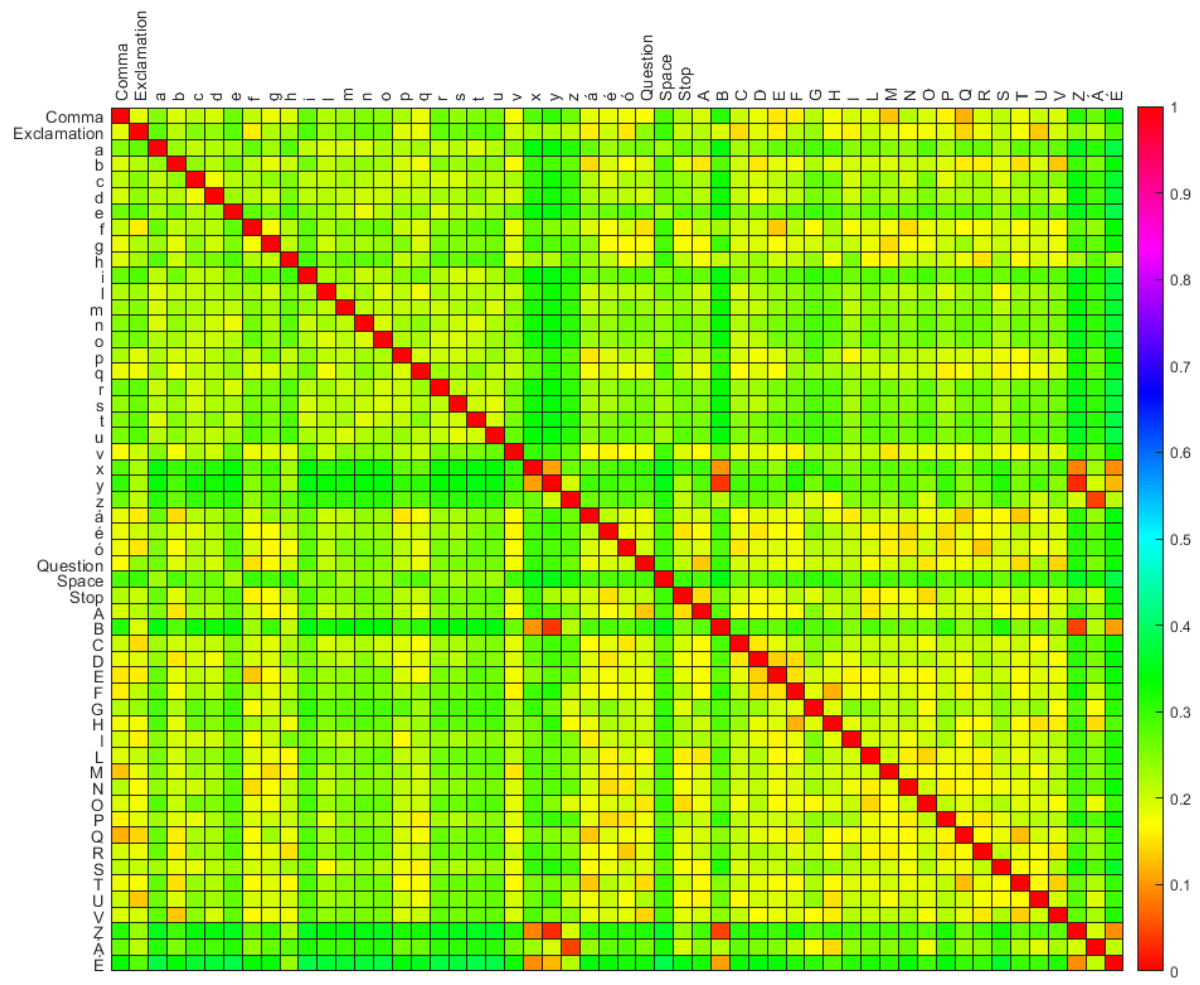} 
        \caption{}
        \label{fig:heatmap_two_full}
    \end{subfigure}
    \begin{subfigure}[t]{0.49\columnwidth}
        \includegraphics[width=1\linewidth]{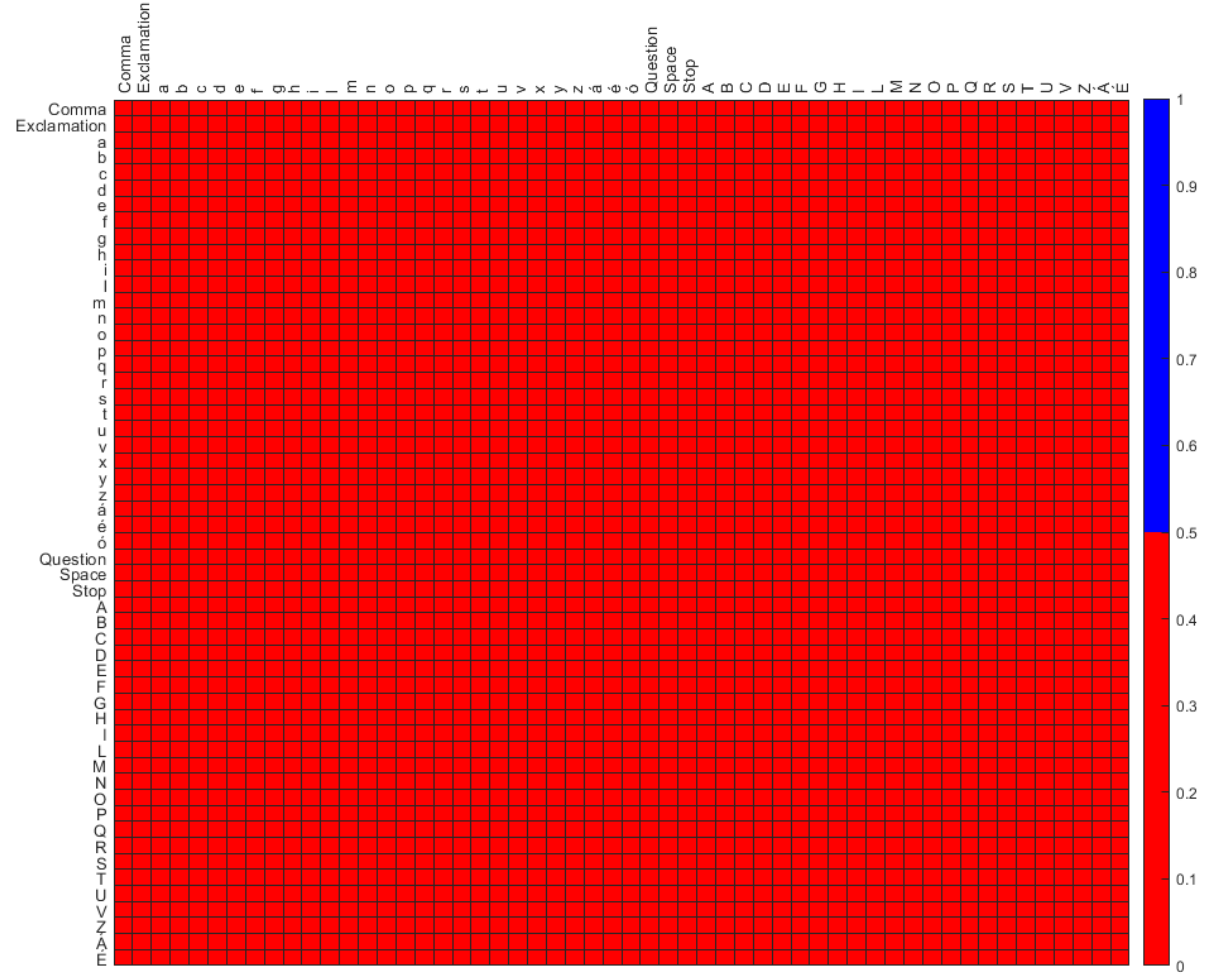} 
        \caption{}
        \label{fig:heatmap_two_thresholded}
    \end{subfigure}
    \caption{Heatmaps of the Bivariate SPIKE-Distances between the Ideal Response Spike Trains of the 54 unique characters of the second stimulus input. On Heatmap b) Red values are below the threshold (0.5)}
    \label{fig:heatmaps_two}
\end{figure}

The heatmaps in figure~\ref{fig:heatmaps_two} for the second stimulus rather starkly confirm the suspicion that prompted this line of enquiry. Of the 54 unique characters in the second stimulus, none of their Ideal Response Spike Trains are separable via the sparseness metric of the novelty search algorithm. In concise terms, the stimulus is not fit for purpose.

\section{Paths Forward}
At this time, in order to progress the Artificial Neural Microcircuit's a building blocks concept, the matter of the effective, efficient generation of those Microcircuits has become the sticking point. Two paths forward present themselves, each of which offer different pros and cons.  

\subsection{Stimulus Optimisation}
The most immediate path would be to pick up the question of an optimal test stimulus. Taking the analysis method employed in the previous section, that being the use of ideal response spike trains \& spike distance measurements, it could be stated that an ideal test stimulus is one that maximises the difference between its corresponding ideal response spike trains. This in turn presents an obvious methodology in the form of the Evolutionary Algorithm: possible stimuli can be generated with minimal issue; code already exists to generate the corresponding Ideal Response Spike Trains and compute the needed matrix of Bivariate SPIKE-Distances; and the maximisation of the values within the distance matrix form the fitness function.

\begin{figure}[H]
    \centering
    \includegraphics[width=0.8\columnwidth]{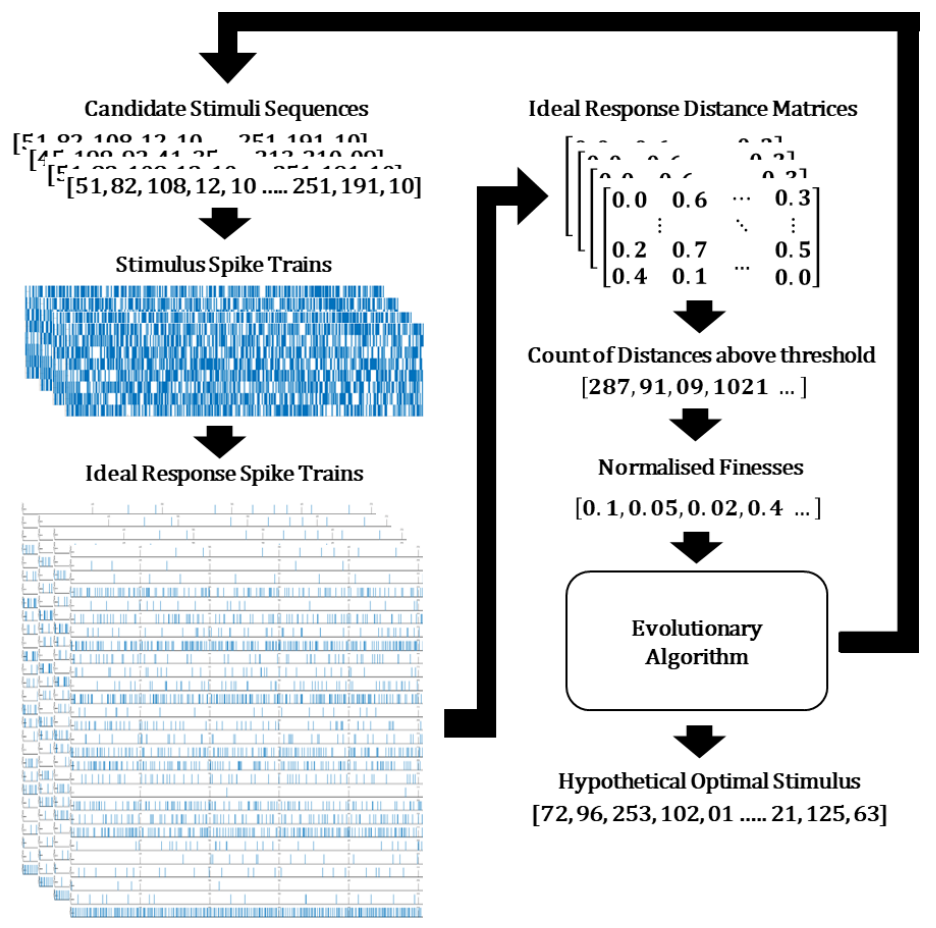}
    \caption{Illustration of the stimulus optimisation algorithm}
    \label{fig:stimulus_optimisation_algorithm}
\end{figure}

To explore the practicality of this stimulus optimisation algorithm, a proof-of-concept experiment was carried out. The population of stimuli consisted of 10 individuals and was iterated over 100 generations. The stimuli consisted of between 255 and 5100 points, drawn from all possible 8- bit values, for a pool of 255 unique input patterns (discounting zeros).

Each new generation was 75\% new individuals, with the other 25\% being drawn from the fittest individuals from the previous generation. New individuals were generated by symmetrical crossover of two parents chosen from the fittest individuals of the previous generation, with the length of the new individual being randomly selected. Mutation had a 15\% chance to replace each pattern in the stimulus with another randomly chosen pattern.

\begin{figure}[H]
    \centering
    \includegraphics[width=0.95\columnwidth]{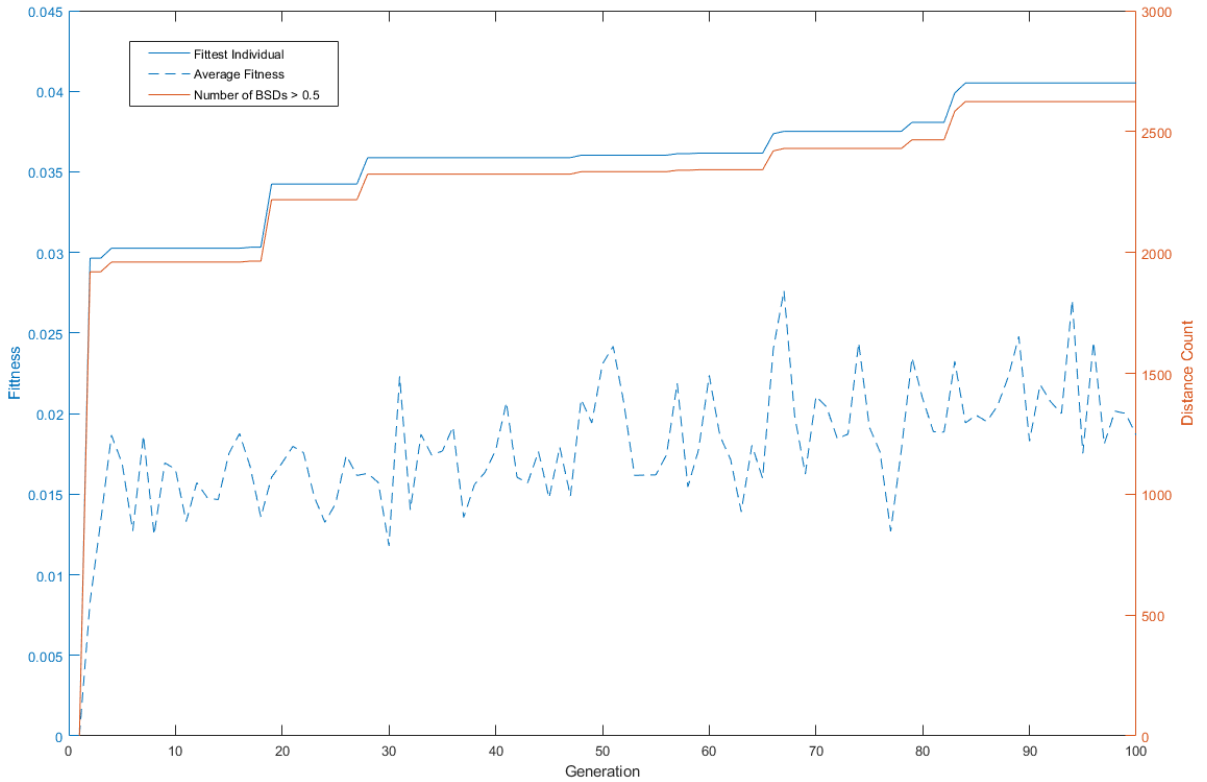}
    \caption{Results of the proof of concept run of the stimulus optimisation algorithm. The left, blue axis is the normalised fitness values; while the right, orange axis is the raw count of distances above the 0.5 threshold}
    \label{fig:stimulus_optimisation_algorithm_results}
\end{figure}

\begin{figure}[H]
    \centering
    \includegraphics[width=0.95\columnwidth]{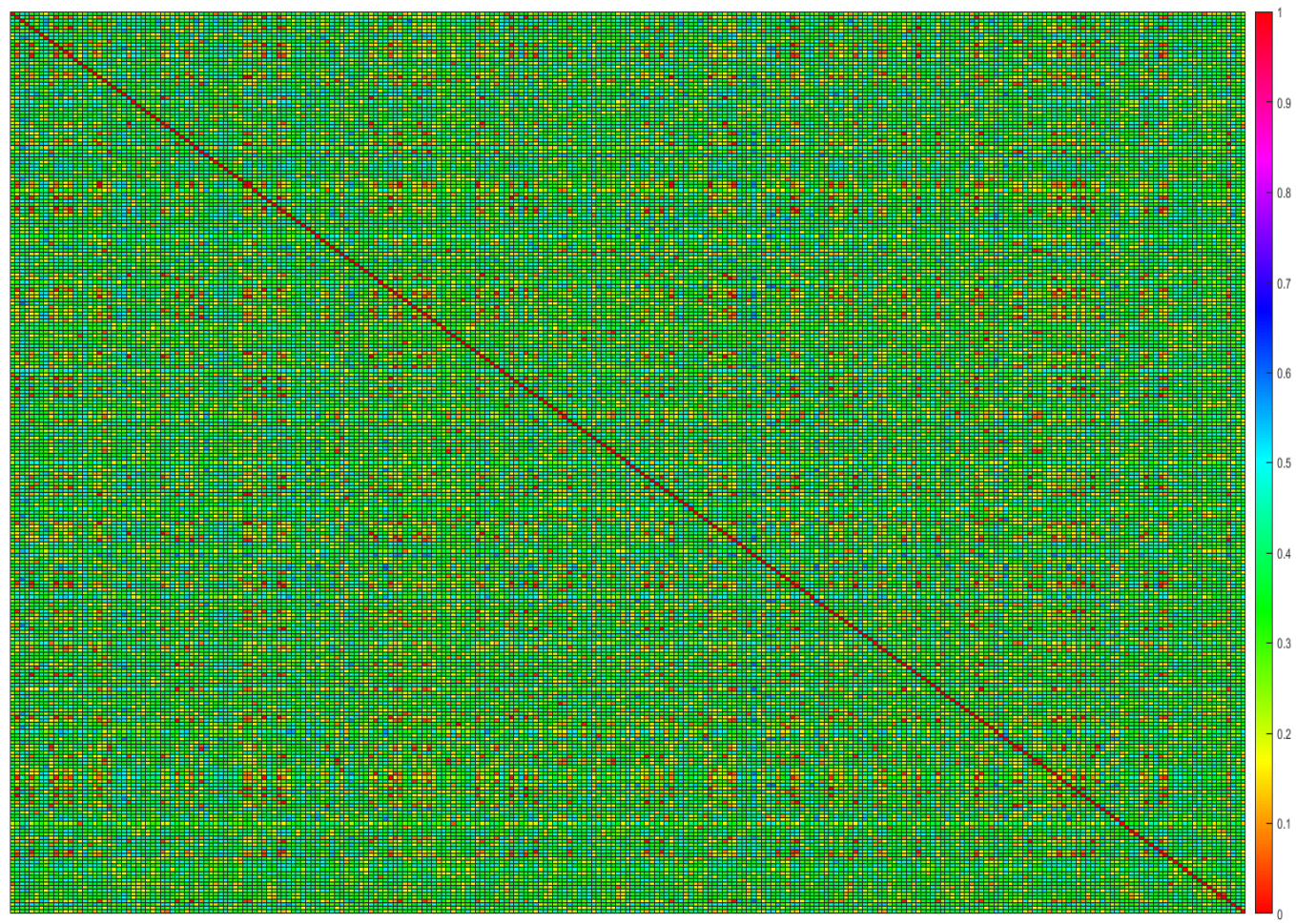}
    \caption{Heatmap of the Bivariate SPIKE-Distances between the Ideal Response Spike Trains of the 255 unique characters of fittest individual produced by the stimulus optimisation experiment.}
    \label{fig:heatmap_three_full}
\end{figure}

\begin{figure}[H]
    \centering
    \includegraphics[width=0.95\columnwidth]{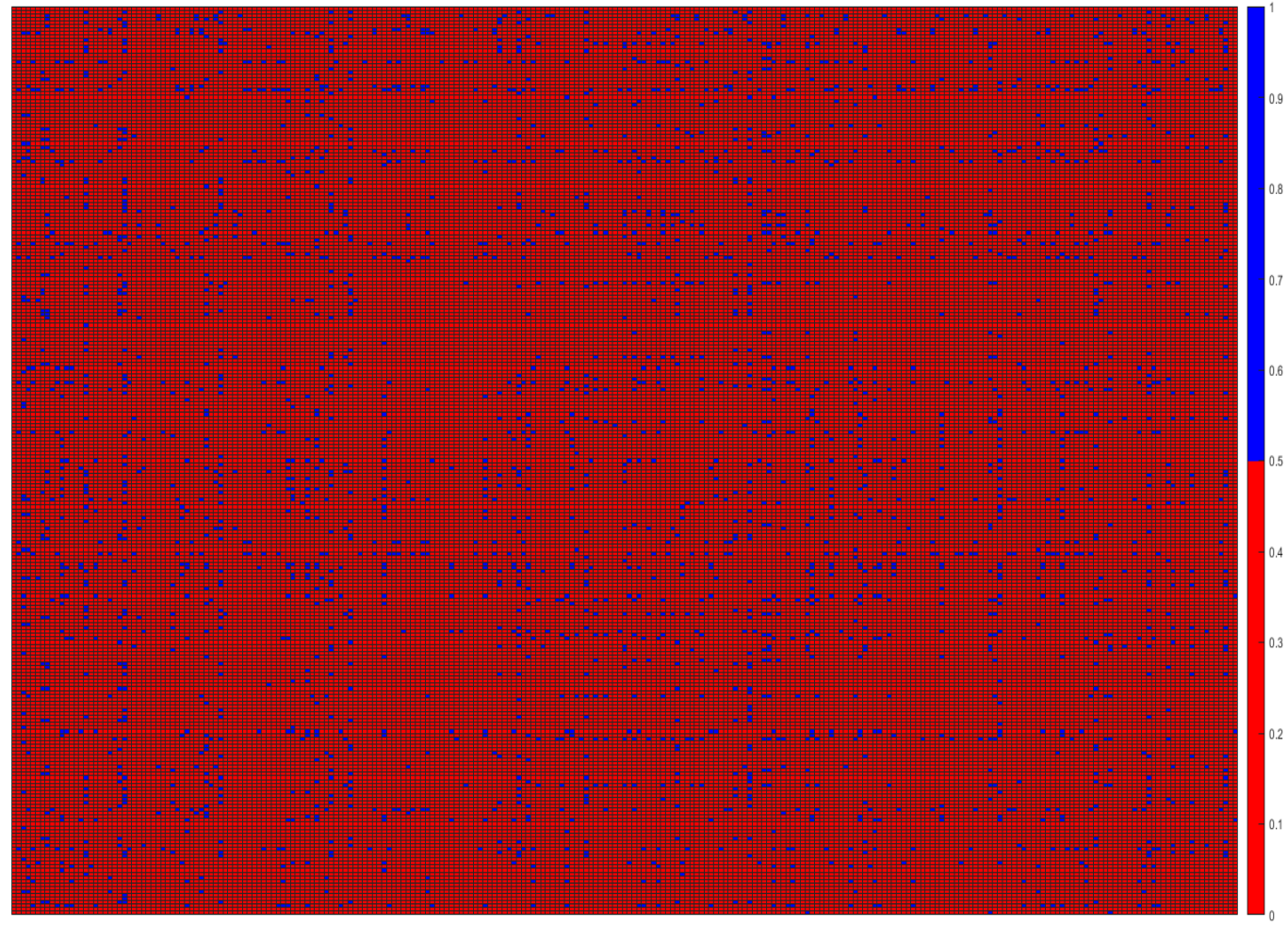}
    \caption{Thresholded heatmap of the Bivariate SPIKE-Distances. Red values are below the threshold (0.5), blue values are above.}
    \label{fig:heatmap_three_thresholded}
\end{figure}

Figure~\ref{fig:stimulus_optimisation_algorithm_results} above shows the generational progress of the proof of concept run. While the increase in fitness was slow, likely due to the small size of the population, the definite upward trend indicates that this approach works. Furthermore, while a direct comparison with the previous text derived stimuli is imperfect, the full and thresholded heatmaps of the fittest individual, shown in figures~\ref{fig:heatmap_three_full} \& ~\ref{fig:heatmap_three_thresholded}, do compare quite positively with those in figures~\ref{fig:heatmaps_two}

However, this methodology does posses a notable drawback: each individual the algorithm assesses needs a 255x255 matrix of Bivariate SPIKE-Distances calculating. Even ignoring self-comparisons (i.e. not computing the distance between b \& a because the distance between a \& b has already been computed) and avoiding re-runs (i.e. not computing a new matrix for an individual that remains unchanged between two generations), this results in a total of 32,640 computations per individual, which given the number of measurements and calculations. 

\subsection{Targeted Evolution}
With the existence of the ideal response spike trains, another possibility that arises is to replace the novelty search based approach with a more targeted search using an evolutionary algorithm. In this methodology, the ideal response spike trains would directly serve as target behaviours, with the goal of the evolutionary algorithm being to minimise the Bivariate SPIKE-Distance between the output of a candidate Microcircuit and a given ideal response spike train.

\begin{figure}[H]
    \centering
    \includegraphics[width=0.95\columnwidth]{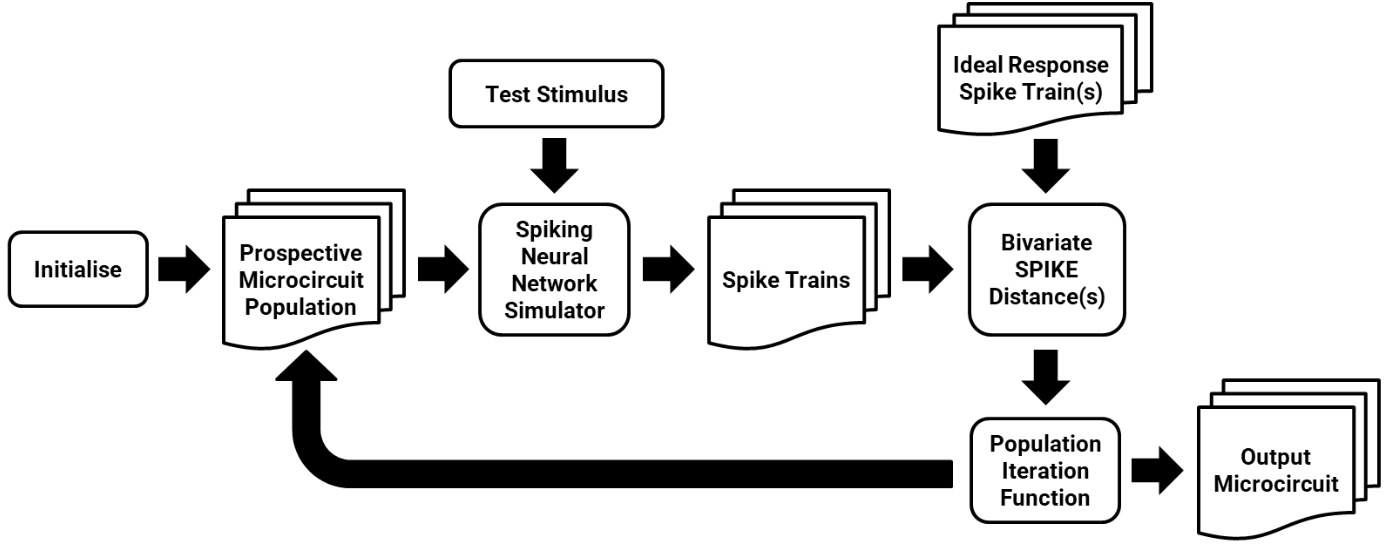}
    \caption{An illustration of the complete Microcircuit evolution methodology, utilising ideal response spike train based fitness.}
    \label{fig:Microcircuit_evolution_methodology}
\end{figure}

The primary positive of this approach is also, paradoxically, its primary downside. The targeting of a specific useful behaviour, or set of behaviours if the algorithm were to be searching using multiple ideal response spike trains in parallel, guarantees that the Microcircuits produced will be useful. This specificity however also removes the possibility of the spontaneous production of useful Microcircuits that is a great strength of the novelty search based approach.      

\section{Conclusions}
In this paper it has been shown that the concept of a catalogue of Artificial Neural Microcircuits, inspired by their biological counterparts, has significant potential as a tool for the creation of more complex Neural Networks. It is straightforward to conceptualise a more expansive version of the Microcircuit catalogue produced during the proof-of-concept experiment parsing UTF-8 text, where Microcircuits could be used as parts of a larger network that takes in handwritten characters, and combined with other kinds of Microcircuit form the interface between the character images and the character recognisers; or in a completely different application where the 8-bit patterns they recognise are some other stimuli of interest.

This modularity also carries with it advantages with respect to implementation in hardware. It is easy to envision how a system built up from Microcircuits could be made only as large as needed, thus avoiding any overhead that might emerge in a more conventional neural network. Furthermore, due to the strongly local nature of the Microcircuits, the amount of longer range signal routing would also be dramatically reduced, further reducing possible hardware complexity.  

The subsequent experiments from Sections 5 \& 6 have shown the current primary barrier to this goal is the refinement of the catalogue generation methodology, specifically with respect to the question of how to ensure the search mechanism provides a rich selection of behaviours that are not just \textit{novel} but \textit{useful}. At this time, the logical option seems to be the pursue something of a combination of the broad searching novelty based approach to produce a rich selection of behaviours, followed by the use of the more targeted evolution based one to refine those behaviours and fill in any gaps.

With this clear path forward, the goal of a usefully expansive catalogue of Microcircuit components and their employment as building blocks for Neuromorphic systems is very much a worthwhile \& obtainable one. 


\bibliographystyle{IEEEtran}
\bibliography{citations}

\begin{thebibliography}{10}
\providecommand{\url}[1]{#1}
\csname url@samestyle\endcsname
\providecommand{\newblock}{\relax}
\providecommand{\bibinfo}[2]{#2}
\providecommand{\BIBentrySTDinterwordspacing}{\spaceskip=0pt\relax}
\providecommand{\BIBentryALTinterwordstretchfactor}{4}
\providecommand{\BIBentryALTinterwordspacing}{\spaceskip=\fontdimen2\font plus
\BIBentryALTinterwordstretchfactor\fontdimen3\font minus
  \fontdimen4\font\relax}
\providecommand{\BIBforeignlanguage}[2]{{%
\expandafter\ifx\csname l@#1\endcsname\relax
\typeout{** WARNING: IEEEtran.bst: No hyphenation pattern has been}%
\typeout{** loaded for the language `#1'. Using the pattern for}%
\typeout{** the default language instead.}%
\else
\language=\csname l@#1\endcsname
\fi
#2}}
\providecommand{\BIBdecl}{\relax}
\BIBdecl

\bibitem{Prieto-2016}
A.~Prieto, B.~Prieto, E.~M. Ortigosa, E.~Ros, F.~Pelayo, J.~Ortega, and
  I.~Rojas, ``Neural networks: an overview of early research, current
  frameworks and new challenges,'' \emph{Neurocomputing}, 2016.

\bibitem{Yamazaki-2022}
K.~Yamazaki, V.-K. Vo-Ho, D.~Bulsara, and N.~Le, ``Spiking neural networks and
  their applications: A review,'' \emph{Brain Sciences}, 2022.

\bibitem{Alzubaidi-2021}
L.~Alzubaidi, J.~Zhang, A.~J. Humaidi, A.~Al-Dujaili, Y.~Duan, O.~Al-Shamma,
  J.~Santamaría, M.~A.~Fadhel, M.~Al-Amidie, and L.~Farhan, ``Review of deep
  learning: concepts, cnn architectures, challenges, applications, future
  directions,'' \emph{Journal of Big Data}, 2021.

\bibitem{Schmidhuber-2015}
J.~Schmidhuber, ``Deep learning in neural networks: An overview,'' \emph{Neural
  Networks}, 2015.

\bibitem{wang-2020}
X.~Wang, X.~Lin, and X.~Dang, ``Supervised learning in spiking neural networks:
  A review of algorithms and evaluations,'' \emph{Neural Networks}, 2020.

\bibitem{Shifei-2013}
D.~Shifei, H.~Li, C.~Su, J.~Yu, and F.~Jin, ``Evolutionary artifical neural
  networks: a review,'' \emph{Artifical Inteligence Review}, 2013.

\bibitem{Stanley-2002}
K.~O. Stanley and R.~Miikkulainen, ``Evolving neural networks through
  augmenting topologies,'' \emph{Evolutionary Computation}, 2002.

\bibitem{Luo-2016}
L.~Luo, \emph{Principles of Neurobiology}.\hskip 1em plus 0.5em minus
  0.4em\relax Garland Science, 2016.

\bibitem{Grillner-2006}
S.~Grillner and A.~M. Graybiel, \emph{Microcircuits: the Interface Between
  Neurons and Global Brain Function}.\hskip 1em plus 0.5em minus 0.4em\relax
  MIT Press, 2006.

\bibitem{Demyanenko-2019}
S.~V. Demyanenko, V.~A. Dzreyan, and A.~Uzdensky, ``Axotomy-induced changes of
  the protein profile in the crayfish ventral cord ganglia,'' \emph{Journal of
  Molecular Neuroscience}, 2019.

\bibitem{Smarandache-Wellmann-2014}
C.~Smarandache-Wellmann and S.~Grätsch, ``Mechanisms of coordination in
  distributed neural circuits: Encoding coordinating information,''
  \emph{Journal of Neuroscience}, 2014.

\bibitem{Schneider-2018}
A.~Schneider, F.~Blumenthal, and C.~Smarandache-Wellmann, ``Adaptive encoding
  of coordinating information in the crayfish central nervous system,''
  \emph{bioRxiv Preprint}, 2018.

\bibitem{Vu-1997}
E.~T. Vu, A.~Berkowitz, and F.~B. Krasne, ``Postexcitatory inhibition of the
  crayfish lateral giant neuron: A mechanism for sensory temporal filtering,''
  \emph{Journal of Neuroscience}, 1997.

\bibitem{Edwards-2017}
D.~Edwards, ``Crayfish escape,'' \emph{Oxford Research Encyclopedia of
  Neuroscience}, 2017.

\bibitem{Byrn-1997}
J.~H. Byrn, \emph{Neuroscience Online: An Electronic Textbook for the
  Neurosciences}.\hskip 1em plus 0.5em minus 0.4em\relax Department of
  Neurobiology and Anatomy - The University of Texas Medical School at Houston,
  1997, ch. Introduction to Neurons and Neuronal Networks.

\bibitem{Bartz-2014}
T.~Bartz-Beielstein, J.~Branke, J.~Mehnen, and O.~Mersmann, ``Evolutionary
  algorithms,'' \emph{WIREs Data Mining and Knowledge Discovery}, 2014.

\bibitem{Lehman-2008}
J.~Lehman and K.~O. Stanley, ``Exploiting open-endedness to solve problems
  through the search for novelty,'' \emph{Proceedings of the Eleventh
  International Conference on Artificial Life}, 2008.

\bibitem{Kreuz-2011}
T.~Kreuz, ``Measures of spike train synchrony,'' \emph{Scholarpedia}, 2011.

\bibitem{Kreuz-2010}
T.~Kreuz, D.~Chicharro, M.~Greschner, and R.~G. Andrzejak, ``Time-resolved and
  time-scale adaptive measures of spike train synchrony,'' \emph{Journal of
  Neuroscience Methods}, 2010.

\end{thebibliography}

\end{document}